\documentclass[letterpaper]{article} 
\usepackage{aaai2026}  

\usepackage{times}  
\usepackage{helvet}  
\usepackage{courier}  
\usepackage[hyphens]{url}  
\usepackage{graphicx} 
\usepackage{natbib}  
\usepackage{caption} 
\usepackage{amsmath}
\usepackage{amssymb}
\usepackage{amsthm}

\newtheorem{proposition}{Proposition}

\usepackage[ruled,lined,longend,fillcomment,linesnumbered,resetcount,titlenotnumbered]{algorithm2e}
\SetKw{InKw}{\textbf{in}}

\DontPrintSemicolon
\SetAlCapSkip{2ex}
\SetSideCommentRight
\SetFillComment

\usepackage{bm}
\usepackage{booktabs}
\usepackage{makecell}
\usepackage{multirow}
\usepackage{chngcntr}

\newcommand{\sectionwithouttoc}[1]{%
  \refstepcounter{section}%
  \sectionmark{#1}%
  \section*{\thesection\quad #1}%
}

\newcommand{\subsectionwithouttoc}[1]{%
  \refstepcounter{subsection}%
  \subsectionmark{#1}%
  \subsection*{\thesubsection\quad #1}%
}

\newcommand{\mbf}[1]{\bm{#1}}

\title{MoETTA: Test-Time Adaptation \\Under Mixed Distribution Shifts with MoE-LayerNorm}

\author{
    Xiao Fan\textsuperscript{\rm 2, \rm 4}\protect\thanks{Work completed during an internship at Tsinghua University. Email: xiaofan140@gmail.com}, Jingyan Jiang\textsuperscript{\rm 1}\protect\thanks{Corresponding author. Email: jiangjingyan@sztu.edu.cn}, Zhaoru Chen\textsuperscript{\rm 3},\\
    Fanding Huang\textsuperscript{\rm 2}, Xiao Chen\textsuperscript{\rm 2}, Qinting Jiang\textsuperscript{\rm 2}, Bowen Zhang\textsuperscript{\rm 1}, Xing Tang\textsuperscript{\rm 1}, Zhi Wang\textsuperscript{\rm 2}
}
\affiliations{
    \textsuperscript{\rm 1}School of Artificial Intelligence, Shenzhen Technology University, Shenzhen, China\\
    \textsuperscript{\rm 2}Shenzhen International Graduate School, Tsinghua University, Shenzhen, China\\
    \textsuperscript{\rm 3}College of Application Technology, Shenzhen University, Shenzhen, China\\
    \textsuperscript{\rm 4}College of Computer Science and Technology, Tongji University, Shanghai, China
}

\begin{document}

\maketitle

\begin{abstract}
Test-Time adaptation (TTA) has proven effective in mitigating performance drops under single-domain distribution shifts by updating model parameters during inference. However, real-world deployments often involve mixed distribution shifts, where test samples are affected by diverse and potentially conflicting domain factors, posing significant challenges even for state-of-the-art TTA methods. A key limitation in existing approaches is their reliance on a unified adaptation path, which fails to account for the fact that optimal gradient directions can vary significantly across different domains. Moreover, current benchmarks focus only on synthetic or homogeneous shifts, failing to capture the complexity of real-world heterogeneous mixed distribution shifts.
To address this, we propose \textbf{MoETTA}, a novel entropy-based TTA framework that integrates the Mixture-of-Experts (MoE) architecture. Rather than enforcing a single parameter update rule for all test samples, MoETTA introduces a set of structurally decoupled experts, enabling adaptation along diverse gradient directions. This design allows the model to better accommodate heterogeneous shifts through flexible and disentangled parameter updates.
To simulate realistic deployment conditions, we introduce two new benchmarks: \textit{potpourri} and \textit{potpourri+}. While classical settings focus solely on synthetic corruptions (i.e., ImageNet-C), potpourri encompasses a broader range of domain shifts—including natural, artistic, and adversarial distortions—capturing more realistic deployment challenges. Additionally, potpourri+ further includes source-domain samples to evaluate robustness against catastrophic forgetting.
Extensive experiments across three mixed distribution shifts settings show that MoETTA consistently outperforms strong baselines, establishing new state-of-the-art performance and highlighting the benefit of modeling multiple adaptation directions via expert-level diversity.
\end{abstract}

\begin{links}
    \link{Code}{https://github.com/AnikiFan/MoETTA}
\end{links}

\sectionwithouttoc{Introduction}
Deep learning has achieved remarkable success. However, it typically relies on the assumption that the source domain, on which the model is trained, shares the same distribution as the target domain, where the model is deployed~\citep{domain_theory}. In practice, this assumption often fails to hold, leading to significant performance degradation under distribution shifts~\citep{wang2024search}.
Test-Time adaptation (TTA)~\citep{Tent,EATA,DeYO,SAR,MGTTA} has emerged as a promising solution to this problem, enabling models to adapt to unseen test distributions without requiring labeled data from the target domain.
TTA methods adapt the model \( f(\cdot; \theta) \) by updating its parameters through the minimization of an unsupervised loss function \( \mathcal{L}(\theta) \), e.g., entropy loss~\citep{Tent}, rotation prediction loss~\citep{gidaris2018unsupervised} and contrastive-based loss~\citep{TTT1}.  

\begin{figure}
  \centering
\includegraphics{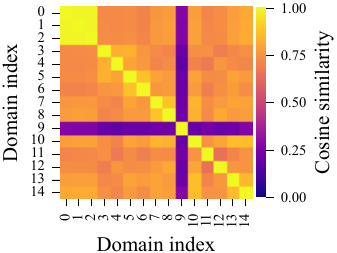}
\caption{
Cosine similarity heatmap of accumulated gradient directions $\theta_i - \theta_{\text{pre}}$, where $\theta_i$ denotes the adapted model parameters on the $i$-th domain of ImageNet-C~\citep{imagenet-c} using Tent~\citep{Tent}, and $\theta_{\text{pre}}$ is the pre-trained model parameters. Each entry at position $(i, j)$ represents the cosine similarity between adaptation directions from domains $i$ and $j$. The average cosine similarity in the lower triangle is 0.69, suggesting substantial variation across domains and highlighting the limitation of using a single adaptation direction under mixed distribution shifts.
}
\label{fig:motivation}
\end{figure}

Early works on TTA, such as Tent~\citep{Tent} and EATA~\citep{EATA}, primarily focused on addressing single-domain distribution shifts, where all test samples originate from the same domain. \citet{SAR} pointed out that existing TTA methods fail under more realistic settings, with mixed distribution shifts as a representative case.
For instance, in real-world deployment scenarios, inference input batches are dynamically constructed from asynchronously arriving, heterogeneous task streams, such as concurrent requests from edge devices, without guarantee of domain consistency~\citep{joint_batch}.
Although recent studies have begun to explore these challenging ``in-the-wild" scenarios, a significant performance gap remains between single-domain and mixed-domain settings. A strong baseline~\citep{MGTTA} reports an accuracy drop from 71.3\% to 65.9\% in the latter setting, which better reflects the challenges of real-world deployment.

To understand this performance gap, we investigate the underlying causes of this limitation. As shown in Fig.~\ref{fig:motivation}, the average cosine similarity between the accumulated gradient directions across the 15 domains of ImageNet-C is only 0.69. This indicates that the gradient directions for different domains are misaligned. Furthermore, assuming that the domain-specific $d$-dimensional parameters at convergence follow a Gaussian distribution, we theoretically show that the expected cosine similarity between the corresponding accumulated gradients, from the pre-trained parameters to the converged ones, converges to $0.5 + \mathcal{O}(1/d)$ (see App.~\ref{app:cosine_similarity_proof}). This result implies the inevitability of aforementioned misalignment. More critically, in some extreme cases, e.g., domain 9 in Fig.~\ref{fig:motivation}, the cosine similarity with other domains can be nearly zero, leading to slow learning and poor generalization~\citep{jacobs1991adaptive}. 
This phenomenon reveals a fundamental limitation of current methods in mixed distribution shifts scenarios: \textit{by enforcing a shared adaptation direction, they fail to accommodate domain-specific gradient signals, which can be inconsistent or even conflicting. }

To address this challenge, we propose \textbf{MoETTA}, a novel test-time adaptation framework that incorporates Mixture-of-Experts (MoE) paradigm~\citep{jacobs1991adaptive, jordan1994hierarchical, shazeer2017outrageously, deepseekmoe} into entropy-based test-time adaptation. Our key \textbf{insight} is that, rather than enforcing a single, unified adaptation path, leveraging a diverse set of experts to represent multiple adaptation solutions within one model is particularly advantageous for mixed distribution shifts. 
Specifically, MoETTA reinterprets the LayerNorm~\citep{layernorm} parameters in Vision Transformer (ViT)~\citep{ViT} as a set of structurally decoupled expert branches. These experts offer distinct parameterizations, enabling the model to accommodate diverse gradient update directions during inference.
Unlike the static domain knowledge vectors used by \citet{CoLa}, our experts are dynamically updated during inference, allowing the model to flexibly absorb distribution-specific signals. What's more, through the routing mechanism, our methods can handle samples in a sample-wise way, which can further decompose diverse gradient directions brought by mixed distribution shifts. 

To reflect more realistic deployment scenarios, we propose two new benchmarks: \textit{potpourri} and \textit{potpourri+}. Compared with the mixed distribution shifts setting proposed by \citet{SAR}, which we refer to as the \emph{classical mixed distribution shifts}, the potpourri setting includes samples from not only ImageNet-C~\citep{imagenet-c} but also ImageNet-R~\citep{imagenet-r}, ImageNet-A~\citep{imagenet-a}, and ImageNet-Sketch~\citep{imagenet-sketch}, forming a heterogeneous mixture of distribution shifts. Beyond synthetic corruptions such as noise, blur, weather, and digital artifacts from ImageNet-C, potpourri introduces additional challenges from natural, artistic, and adversarial distortions, offering a more comprehensive testbed.

We further propose potpourri+, which augments potpourri with ImageNet validation samples to evaluate methods' resilience to catastrophic forgetting when simultaneously handling in-distribution (ID) and out-of-distribution (OOD) data—a common challenge in practical applications.
Our main contributions are summarized as follows:

\begin{itemize}
\item We identify a key limitation of existing TTA methods under mixed distribution shifts and address it with MoETTA, an entropy-based method that leverages expert routing to model diverse adaptation directions.
\item We propose two novel evaluation settings, i.e., potpourri and potpourri+, which better simulate realistic deployment conditions involving mixed distribution shifts.
\item MoETTA achieves state-of-the-art performance on both existing mixed distribution shift benchmarks and the newly proposed potpourri and potpourri+ settings.
\end{itemize}
\sectionwithouttoc{Related Work}
\paragraph{Entropy-based Test-Time Adaptation.}
Entropy-based Test-Time Adaptation~\citep{Tent,EATA,SAR,DeYO,MGTTA,COTTA,BECoTTA} refers to a class of TTA methods that enhance the robustness of pre-trained models under distribution shifts by minimizing prediction entropy~\citep{entropy} during test-time adaptation.

Tent~\citep{Tent}, EATA~\citep{EATA}, SAR~\citep{SAR}, and DeYO~\citep{DeYO} adapt only the affine parameters of normalization layers. Among them, EATA, SAR, and DeYO improve the reliability of gradient signals through selective sample filtering, while SAR further incorporates Sharpness-Aware Minimization (SAM)~\citep{SAM} to enhance generalization under severe shifts. MGTTA~\citep{MGTTA} introduces a meta-gradient generator (MGG) to guide affine parameter updates.

While our method also belongs to the family of entropy-minimization-based approaches, it differs fundamentally in its architectural flexibility and update strategy.
Some recent works have also explored structural modifications to enhance test-time adaptation. Notably, BECoTTA~\citep{BECoTTA} proposes a MoE framework by incorporating domain-specific low-rank modules (LoRA)~\citep{hu2022lora} as experts, aiming to address the challenges of Continual Test-Time Adaptation, where the dominant domain may gradually evolve over time. In contrast, our approach adopts LayerNorm as the expert unit and specifically targets the more challenging setting of mixed distribution shifts. 

\begin{figure*}[!ht]
  \centering
\includegraphics[width=\linewidth]{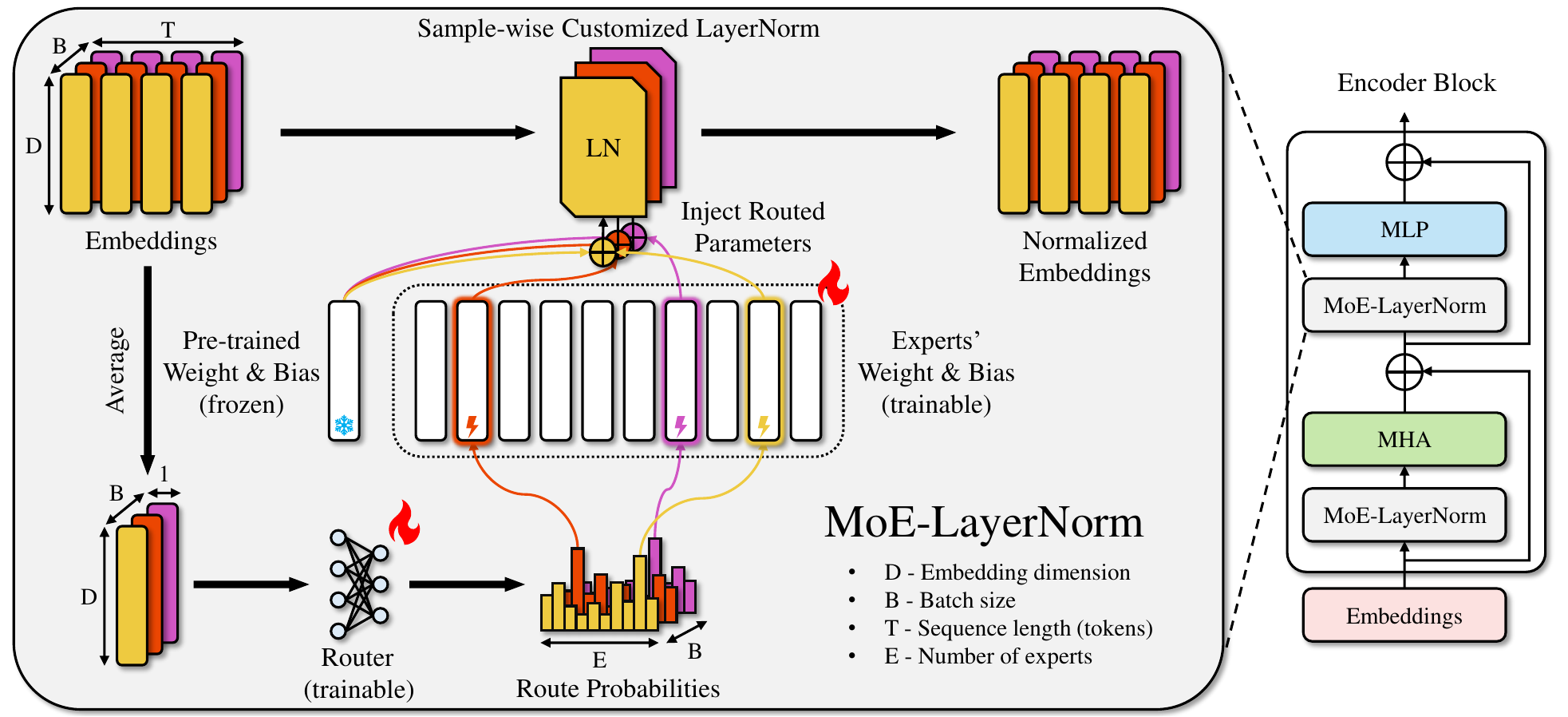}
\captionof{figure}{\textit{Method Overview}. We replace the LayerNorm modules in the encoder blocks of a Vision Transformer (ViT)~\citep{ViT} with our proposed MoE-LayerNorm. Colors of the embeddings and their routed components are used illustratively to suggest that the samples originate from different domains. For each input embedding, we first compute the mean across the token (sequence) dimension. This averaged vector is fed into a router to obtain routing probabilities, and the expert with the highest probability is selected. Its parameters are then added to the frozen pre-trained LayerNorm parameters to form a sample-specific LayerNorm. Finally, each token embedding is normalized using this customized LayerNorm. A PyTorch-style pseudo code of the forward pass for MoE-LayerNorm can be found in App.~\ref{app:pseudo_code}.}
\label{fig:pipeline}
\end{figure*}

\paragraph{Mixture of Experts.}
Mixture of Experts was initially proposed for its ability to handle diverse subtasks by routing inputs to multiple expert networks through a gating mechanism~\citep{jacobs1991adaptive}. In recent years, MoE has gained significant traction in natural language processing (NLP)~\citep{shazeer2017outrageously,deepseekmoe,DeepSeekV3,switch_transformer,mixtral,MoE}, as it enables large-scale model capacity while maintaining a limited computational footprint during inference. For example, \citet{switch_transformer} proposed a Mixture-of-Experts layer for natural language processing, where multiple feed-forward networks with distinct parameters are treated as experts. During inference, the top-1 experts are selected based on the output of a router, which consists of a single fully connected layer.

Similar to the Deep Mixture of Experts (DMoE) framework proposed by \citet{eigen2014learningfactoredrepresentationsdeep}, our method introduces an MoE structure at every layer of the model to support diverse computation pathways. However, a key distinction lies in how the experts are implemented: while DMoE uses full linear transformations followed by nonlinearities as experts, we adopt LayerNorm instances as lightweight experts. This design choice significantly reduces the computational overhead, as only the affine parameters of LayerNorm are adapted per expert. As a result, \textit{our approach preserves expert diversity while maintaining high efficiency, making it well-suited for test-time adaptation}.
\sectionwithouttoc{Methodology}

We begin this section by formally defining the mixed distribution shifts setting. We then introduce the core component of our method, the MoE-LayerNorm module, as illustrated in Fig.~\ref{fig:pipeline}. Finally, we detail the overall adaptation procedure.

\subsectionwithouttoc{Problem Statement}
Consider a pre-trained model \( f(\cdot; \theta) \), where \( \theta \) is learned from a labeled dataset \( \{(\mbf x_n, y_n)\}_{n=1}^{N} \) drawn from a source distribution \( \mathcal{P}(\mbf x, y) \). Once deployed, without access to ground-truth labels $y$, the model processes a stream of input sample batches \( \mathcal{B}_0, \mathcal{B}_1, \ldots \), each of size \( B \), drawn from a target distribution that typically differs from \( \mathcal{P}(\mbf x, y) \), resulting in distribution shifts that can significantly degrade performance. 
In the conventional single-domain setting, all test samples---both within and across batches---are assumed to come from a single target distribution \( \mathcal{Q}(\mbf x, y) \), with \( \mathcal{Q} \ne \mathcal{P} \). 
In contrast, the more realistic mixed-domain setting allows each batch to contain samples from multiple target distributions \( \mathcal{Q}_1(\mbf x, y), \mathcal{Q}_2(\mbf x, y), \ldots \), capturing diverse, domain-specific shifts encountered during deployment.

\subsectionwithouttoc{MoE-LayerNorm}
\label{MoE-LayerNorm}

\paragraph{Key Design Choices.}
Our method adapts MoE to the test-time adaptation setting through several key design choices:
\begin{enumerate}
    \item To minimize computational overhead—a critical requirement for TTA—we designate the LayerNorm parameters as experts instead of employing costly feed-forward networks. During adaptation, only the router and expert-specific LayerNorm parameters are updated, ensuring lightweight and efficient optimization. 
    \item To promote diversity across experts while maintaining efficiency, we follow the top-1 routing strategy adopted by \citet{switch_transformer}, activating only one expert per sample. This design also avoids the need to merge multiple experts' output, thereby preventing parameter interference and preserving their individuality.
    \item Given the limited number of samples at test time, we construct the effective LayerNorm parameters used for normalizing each sample as the sum of the corresponding expert’s parameters (initialized as zeros) selected by the router and the frozen pre-trained LayerNorm parameters, which we refer to as the \textit{shared expert}. 
    The inclusion of the shared expert provides domain-invariant knowledge, such as semantic representations learned from ID samples. In addition, it serves as a common foundation across all samples, which helps reduce redundancy among experts and encourages them to develop complementary adaptation behaviors~\citep{deepseekmoe}.
\end{enumerate}

\paragraph{Routing Mechanism.}

Our router is a linear projector that takes the mean sample embedding over the token dimension as input and outputs a vector whose dimension equals the number of experts. It is initialized with Xavier initialization~\citep{xavier} at the start of adaptation.


Following \citet{switch_transformer}, in addition to task loss, we update routers with a differentiable load balancing loss.
Given \(N\) experts indexed from \(1\) to \(N\) and the \(t\)-th batch \(\mathcal{B}_t\), the auxiliary loss is computed as the scaled dot-product between two vectors \(\mbf F\) and \(\mbf P\):
\begin{equation}
\mathcal{L}_{\text{load balancing}} = N\times \sum_{i=1}^N \mbf F_i \times \mbf P_i.
\label{eq:load_balancing}
\end{equation}
Here, \(\mbf F_i\) denotes the fraction of samples in \(\mathcal{B}_t\) routed to expert \(i\), and \(\mbf P_i\) represents the average router-assigned probability for expert \(i\) w.r.t. samples in \(\mathcal{B}_t\):
{\fontsize{9pt}{11pt}\selectfont
\begin{equation}
\mbf F_i = \frac{1}{|\mathcal{B}_t|} \sum_{\mbf x \in \mathcal{B}_t} \mathbb{I}_{\{\arg\max_k \mbf p_k(\mbf x) = i\}},
\mbf P_i = \frac{1}{|\mathcal{B}_t|} \sum_{\mbf x \in \mathcal{B}_t} \mbf p_i(\mbf x),
\end{equation}
}
where $\mbf p(\mbf x)$ denotes the full routing probability vector produced by the router for sample $\mbf x$, while $\mbf p_i(\mbf x)$ refers to its $i$-th component.

The aforementioned loss primarily encourages balanced expert utilization across samples, as it reaches its minimum value $1$ when all elements in $\mbf F$ are equal. A theoretical proof is provided in App.~\ref{app:load_balancing}.

However, in the test-time adaptation setting, balanced routing alone is insufficient. We also aim for the router to establish a consistent mapping that aligns inputs with appropriate experts based on shared adaptation patterns. In other words, it is desirable for samples with similar characteristics to be routed to the same expert, enabling each expert to adapt along distinct update directions rather than responding to a heterogeneous mixture of inputs.

To support such specialization, we adopt the following strategy: during the forward pass, only the expert corresponding to the maximum routing probability \( p \) is activated. To ensure that the router remains trainable under the entropy-based task loss, we apply a gradient-preserving trick by scaling the selected expert output with \( p / p.\texttt{detach}() \). This operation does not affect the forward computation but allows gradients to flow from the entropy loss into the router. As a result, the routing decisions are optimized jointly with model predictions, encouraging experts to develop diverse and input-sensitive behaviors over time.

For the $t$-th batch $\mathcal{B}_t$, to maintain a balance between the load balancing loss $\mathcal{L}_{\text{load balancing}}$ and the entropy loss during adaptation, we scale $\mathcal{L}_{\text{load balancing}}$  using a dynamic weight $\alpha_t$, defined as:
\begin{equation}
\alpha_t = 
\begin{cases}
\lambda \times E_{\text{avg}}^0, & t = 0 \\
\alpha_{t-1} \times \dfrac{E_{\text{avg}}^t}{E_{\text{avg}}^{t-1}}, & t \geq 1
\end{cases},
\label{eq:lb_alpha}
\end{equation}
where $\lambda$ is a hyper-parameter that controls the relative importance of the load balancing loss w.r.t. the entropy loss, and $E^t_{avg}$ denotes the average entropy loss of all test samples up to the $t$-th batch.

\subsectionwithouttoc{Overall Procedure}
Given an image $\mbf x\in\mathbb{R}^{H\times W\times C}$, we divide this image into $N=HW/P^2$ patches $\{\mbf x_p^i\}_{i=1}^N$, each with shape $P\times P\times C$. Next, we apply $D$ filters with the same shape to each patch to create a patch embedding $\tilde{\mbf z}_0^i\in\mathbb{R}^{D\times 1}(i=1,\dots,N)$. We then concatenate the patch embedding with the class embedding $\mbf x_{\mathrm{class}}=\tilde{\mbf z}_0^0$. Finally, we add the position encoding $\mbf E_{\mathrm{pos}}\in \mathbb{R}^{D\times(N+1)}$ to obtain the final embedding at the layer 0. With a $L$-layers ViT, the forward pass of MoETTA can be summarized as Eqs.~\eqref{eq:embedding}--\eqref{eq:final_rep}, where MSA stands for Multi-head Self-Attention~\citep{attention} and MLP stands for Multi-Layer Perceptron.
{\small
\begin{align}
    \mbf z_{0} &=
      \bigl[\,\mbf{x}_{\mathrm{class}};\,
              \mbf{x}^{1}_{p}\mbf E;\,
              \mbf{x}^{2}_{p}\mbf E;\,
              \ldots;\,
              \mbf{x}^{N}_{p}\mbf E\bigr]
      + \mbf E_{\mathrm{pos}},
      \label{eq:embedding} \\
    \mbf z'_{\ell} &=
      \operatorname{MSA}\!\bigl(
        \operatorname{MoE\text{-}LayerNorm}(\mbf z_{\ell-1})
      \bigr)
      + \mbf z_{\ell-1},
      \label{eq:msa_apply} \\
    \mbf z_{\ell} &=
      \operatorname{MLP}\!\bigl(
        \operatorname{MoE\text{-}LayerNorm}(\mbf z'_{\ell})
      \bigr)
      + \mbf z'_{\ell},
      \label{eq:mlp_apply} \\
\mbf{p}\!\left(y \middle| \mbf{x}\right) &= \operatorname{Softmax}\left( \operatorname{MLP}\left( \operatorname{MoE\text{-}LayerNorm}(\mbf z^{0}_{L}) \right) \right),
      \label{eq:final_rep}
\end{align}
}
where $\mbf E\!\in\!\mathbb R^{D\times(P^{2}C)},\mbf E_{\mathrm{pos}}\!\in\!\mathbb R^{D\times(N+1)}$ and $\ell = 1,\dots,L$.

Inspired by \citet{CEMA} and \citet{EATA}, we filter out samples with high entropy loss using a dynamic threshold and re-weight remaining samples by their entropy loss. Specifically, for the $t$-th batch $\mathcal{B}_t$, once the forward pass is finished, we calculate the sample-selection threshold \(E_{\max}^{t}\) using
\begin{equation}
E_{max}^t = 
\begin{cases}
E_{avg}^0, & t = 0 \\
E_{max}^{t-1}\times\frac{E_{avg}^t}{E_{avg}^{t-1}}, & t \geq 1
\end{cases},
\end{equation}
We then back-propagate the following multi-term loss:
{\fontsize{9pt}{11pt}\selectfont
\begin{equation}
 \frac{1}{\lvert\mathcal{S}_{t}\rvert}
  \sum_{\mbf x\in\mathcal{S}_{t}}
    \underbrace{\exp{\left[\mathrm{E}_0-\mathrm{Ent}(\mbf x)\right]}}_{\text{Entropy re-weighting}}\underbrace{\mathrm{Ent}(\mbf x)}_{\text{Entropy loss}}
  +\alpha_t\sum_{i=1}^{M}\mathcal{L}^i_{\text{load balancing}},
  \label{eq:overall_loss}
\end{equation}
}
where \(\mathrm{Ent}(\mbf x)\) denotes the entropy of the posterior probability distribution $\mbf{p}\!\left(y \middle| \mbf{x}\right)$ in Eq.~\eqref{eq:final_rep}, \(\alpha_t\), defined by Eq.~\eqref{eq:lb_alpha}, is the trade-off coefficient that balances the entropy-minimization term against the load balancing loss, $M$ is the number of MoE-LayerNorm and $\mathcal{L}^i_{\text{load balancing}}$ is the load balancing loss for the router corresponding to the $i$-th MoE-LayerNorm. The set of samples that participate in the entropy term is
\begin{equation}
  \mathcal{S}_{t}
  =\bigl\{\mbf x\;\bigl|\;
    \mathrm{Ent}(\mbf x)<E_{\max}^{t} \land \mbf x\in \mathcal{B}_t
  \bigr\},\label{eq:St}
\end{equation}
i.e., the “reliable” samples whose entropy falls below the current threshold.  Given that there are only two degrees of freedom among the learning rate, $\lambda$ in Eq.~\eqref{eq:lb_alpha}, and $\mathrm{E}_0$ in Eq.~\eqref{eq:overall_loss}, we fix $\mathrm{E}_0$ as a constant and tune the others as hyper-parameters throughout this work. 


To clarify, among all symbols in Eqs.~\eqref{eq:load_balancing}--\eqref{eq:St}, only two need to be manually specified: $\lambda$ and $M$, determined by the chosen MoE-LayerNorm replacement strategy. In Sec.~\ref{subsec:hyper}, we further demonstrate MoETTA's robustness to $\lambda$ and provide practical guidelines for the replacement strategy. Overall, MoETTA remains flexible and easy to tune. The pseudocode of the entire procedure is presented in App.~\ref{app:pseudo_code}.

\sectionwithouttoc{\textit{Potpourri} and \textit{Potpourri+} Benchmarks}
\label{benckmark}
\begin{figure}

\centering
\includegraphics{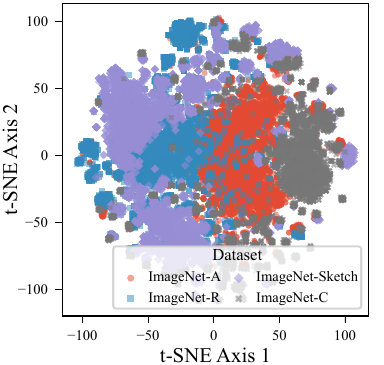}
\caption{
t-SNE~\citep{tsne} projection of CLS token embeddings $\mbf z^{0}_{L}$ in Eq.~\eqref{eq:final_rep}, from ViT-B/16 on 7,500 samples per dataset from ImageNet-A, -R, -Sketch, and -C.
While classical mixed distribution shifts (ImageNet-C only) occupy a relatively narrow region (gray), our proposed potpourri benchmark introduces greater semantic and stylistic diversity by incorporating three additional variants.
This results in a more heterogeneous and realistic evaluation setting for TTA.
}
\label{fig:dataset}
\end{figure}

Since \citet{SAR} highlighted the difficulty of addressing mixed distribution shifts and adopted a mixture of ImageNet-C as the evaluation setting, this benchmark has become the \textit{de facto} standard for assessing the robustness of TTA methods under mixed-domain scenarios~\citep{DeYO,use_mix_scenario_1,MGTTA,EATA-C}. While ImageNet-C contains 15 corruption types grouped into four categories---noise, blur, weather, and digital---it still falls short in capturing the full diversity and severity of real-world distribution shifts. For example, natural renditions can cause substantial misalignment in texture and local image statistics~\citep{imagenet-r}, and spurious background changes may appear even when the object of interest remains visually consistent~\citep{spurious_distribution_shift}.

To close this gap, we propose the potpourri benchmark, which comprises a heterogeneous mixture of samples from ImageNet-C~\citep{imagenet-c}, ImageNet-R~\citep{imagenet-r}, ImageNet-Sketch~\citep{imagenet-sketch}, and ImageNet-A~\citep{imagenet-a}. ImageNet-R evaluates robustness against stylistic renditions, while ImageNet-A and ImageNet-Sketch probe vulnerability to spurious correlations and semantic abstraction. As shown in Fig.~\ref{fig:dataset}, these datasets complement ImageNet-C and together span a wider and more diverse set of domain shifts. This composition enables a more comprehensive and realistic evaluation of TTA methods.

In addition, test-time data streams may occasionally include samples from the source distribution. However, TTA models often suffer from catastrophic forgetting on such ID samples after being adapted to OOD samples~\citep{EATA}. To address this, we introduce potpourri+, an extension of potpourri that includes validation samples from the original ImageNet dataset. It provides a more faithful assessment of a method's ability to balance adaptation and retention during mixed-domain deployment.

\sectionwithouttoc{Experiments}
\label{sec:experiment}
\subsectionwithouttoc{Experimental Protocol}
We evaluate our method under three settings: the classical mixed distribution shifts, and the proposed potpourri and potpourri+ benchmarks. Details of the related datasets are provided in App.~\ref{app:dataset}. All evaluation settings used in this section consist of ImageNet-C at corruption severity level 5, with a batch size of 64. In this work, all pre-trained models are obtained from timm~\citep{timm} library. Unless otherwise stated, all experiments use ViT-B/16.

We compare MoETTA with the following state-of-the-art test-time adaptation methods: Tent~\citep{Tent}, EATA~\citep{EATA}, SAR~\citep{SAR}, DeYO~\citep{DeYO}, CoTTA~\citep{COTTA}, MGTTA~\citep{MGTTA}, and BECoTTA~\citep{BECoTTA}. Additional implementation details and baseline configurations are provided in App.~\ref{app:implementaion_details}.



\subsectionwithouttoc{Robustness to Mixed Distribution Shifts}
\begin{table*}[t]
  \centering
  {\fontsize{9pt}{11pt}\selectfont
  \begin{tabular}{>{\centering\arraybackslash}c c cccccccc c}
    \toprule
    Model & Setting & Noadapt & Tent & EATA & CoTTA & SAR & DeYO & MGTTA & BECoTTA & Ours \\
    \midrule
    \multirow{3}{*}{\makecell{ViT \\ -B/16}} 
        & Classical & $55.52$ & $63.20_{0.08}$ & $64.28_{0.09}$ & $60.53_{0.57}$ & $60.76_{0.04}$ & $63.97_{0.04}$ & $\underline{66.20}_{0.01}$ & $61.57_{0.08}$ & $\textbf{67.20}_{0.03}$ \\
        & Pot. & $54.18$ & $60.99_{0.05}$ & $61.99_{0.11}$ & $59.67_{1.21}$ & $58.71_{0.03}$ & $61.66_{0.02}$ & $\underline{62.98}_{0.26}$ & $59.08_{0.86}$ & $\textbf{65.12}_{0.08}$ \\
        & Pot.+ & $55.92$ & $62.28_{0.03}$ & $63.17_{0.06}$ & $59.26_{0.68}$ & $59.99_{0.07}$ & $62.90_{0.03}$ & $\underline{64.35}_{0.07}$ & $58.87_{3.23}$ & $\textbf{66.15}_{0.06}$ \\
    \midrule
        \multirow{3}{*}{\makecell{Conv.\\ -B}} 
        & Classical & $54.81$ & $58.88_{0.06}$ & $\underline{64.50}_{0.06}$ & $59.65_{0.04}$ & $61.67_{2.65}$ & $64.32_{0.03}$ & - &$50.16_{7.96}$ & $\textbf{67.40}_{0.02}$ \\
        & Pot. & $53.91$ &  $58.23_{0.05}$ & $\underline{62.69}_{0.07}$ & $58.57_{0.00}$ & $61.16_{0.41}$ & $62.46_{0.07}$ & - & $28.28_{20.54}$ & $\textbf{65.70}_{0.05}$ \\ 
        & Pot.+ & $55.69$ & $59.69_{0.04}$ & $\underline{63.94}_{0.07}$ & $60.02_{0.02}$ & $62.72_{0.10}$ & $63.57_{0.06}$ & - & $48.92_{9.35}$ & $\textbf{66.68}_{0.07}$ \\
    \bottomrule
  \end{tabular}
    }
    \caption{
    Accuracy comparison (\%, $\uparrow$) under different mixed distribution settings. Conv. stands for ConvNeXt. Noadapt refers to the evaluation of the model without adaptation. Results for MGTTA applied to ConvNeXt are not reported, since pre-training MGG for architectures with multiple normalization dimensions is non-trivial. For compactness, the notation $a_{b}$ denotes $a \pm b$, where $b$ represents the standard deviation calculated from the results of three different random seeds. The best performance is highlighted in \textbf{bold} and the second best is indicated by \underline{underlining}. This convention is followed in all subsequent tables.
  }
    \label{tab:mixed_shift_benchmark}
\end{table*}
We evaluate MoETTA on three mixed distribution shift benchmarks using both Transformer-based models (ViT~\citep{ViT}) and convolution-based models (ConvNeXt~\citep{liu2022convnet}). As shown in Tab.~\ref{tab:mixed_shift_benchmark}, MoETTA achieves \textit{state-of-the-art performance across all six settings}, demonstrating its effectiveness in improving the robustness of diverse architectures under mixed distribution shifts.
To assess scalability, we further apply MoETTA to ViT-L/16 (304M parameters) in addition to ViT-B/16 (86M parameters); details are provided in App.~\ref{app:scalability}.


Furthermore, \textit{MoETTA achieves these results without requiring any additional samples for pretraining or the calculation of statistical information}. In contrast, MGTTA, a strong baseline, relies on both extra OOD and ID samples. 

\subsection{Computation Efficiency}
\begin{table}[htbp]
\centering
{\fontsize{9pt}{11pt}\selectfont
\begin{tabular}{lcccc}
\toprule
Method & \makecell{\#Act. Params\\per sample}  & \#Fwd & \#Bwd &  \makecell{Used\\time} \\
\midrule
Noadapt         & 0  & 100\% & 0\% & 100\% \\ 
Tent & 0.04M  & 100\% & 100\% & 226\% \\ 
EATA & 0.04M  & 100\% & 80\% & 239\% \\ 
SAR   & 0.03M & 199\% & 175\% & 440\% \\ 
DeYO   & 0.04M & 196\% & 53\% & 317\% \\ 
CoTTA & 86.42M  & 199\% & 100\% & 798\% \\ 
MGTTA & 2.80M  & 100\% & 100\% & 227\% \\ 
BECoTTA & 0.13M & 100\% & 86\% & 334\%\\ 
Ours & 0.23M & 100\% & 76\% & 247\%  \\ 
\bottomrule
\end{tabular}
 \caption{Computation efficiency comparison, showing per-sample activated parameter count; forward and backward propagation counts (as a percentage of total number of samples); and computation time\textit{ relative to the Noadapt baseline}. All statistics are measured under potpourri+ setting.}
  \label{tab:computation_efficiency}
}
\end{table}

As shown in Tab.~\ref{tab:computation_efficiency}, MoETTA incurs modest overhead, its runtime is 247\% of the model without adaptation, only slightly above the overheads of Tent, EATA, and MGTTA.

\subsectionwithouttoc{Ablation Study}

\begin{table}[ht]
\centering
\fontsize{9pt}{11pt}\selectfont
\begin{tabular}{lcccc}
\toprule
 & Classical & Pot. & Pot.+ & Avg. \\
\midrule
\textbf{Full method} & \textbf{67.25} & \textbf{65.14} & \textbf{66.21} & \textbf{66.20} \\
\midrule
\multicolumn{5}{c}{\textbf{Loss Components}} \\
\midrule
\textit{w/o} Sample selection              & \underline{67.04} & \underline{64.01} & 57.61 & \underline{62.89} \\
\textit{w/o} Entropy re-weight             & 62.86 & 60.51 & \underline{61.79} & 61.72 \\
\textit{w/o} $\mathcal{L}_{\text{load balancing}}$ & 26.27 & 16.27 & 21.29 & 21.28 \\
\midrule
\multicolumn{5}{c}{\textbf{MoE Architecture}} \\
\midrule
\textit{w/o} Grad to router                & \underline{65.17} & \underline{62.80} & \underline{63.92} & \underline{63.96} \\
\textit{w/o} Sample-wise router            & 28.69 & 28.60 & 24.96 & 27.42 \\
\textit{w/o} MoE-LayerNorm                 & 22.38 & 17.94 & 26.93 & 22.42 \\
\textit{w/o} Layer-wise router             & 17.40 & 27.09 & 15.18 & 19.89 \\
\bottomrule
\end{tabular}
\caption{Ablation study. The first group evaluates the effect of different loss components (Eq.~\eqref{eq:overall_loss}), and the second group evaluates the effect of MoE architectural choices. Accuracy (\%, $\uparrow$) is reported under classical mixed distribution shifts.}
\label{tab:ablation_combined}
\end{table}

\paragraph{Ablation Study on Loss Components in Eq.~\eqref{eq:overall_loss}.}
Removing either the sample selection mechanism, i.e., Eq.~\eqref{eq:St}, or the entropy re-weighting part in Eq.~\eqref{eq:overall_loss} leads to moderate performance drops, showing their effectiveness in filtering reliable samples and emphasizing confident predictions. In contrast, removing the load balancing loss leads to a dramatic degradation across all settings, confirming its essential role in maintaining routing stability and preventing collapse.
\paragraph{Ablation Study on MoE Architecture.}
Disabling the gradient flow from the entropy loss to the router reduces the model’s ability to learn informative routing strategies, resulting in a notable performance drop. Disabling sample-wise router causes the same expert to be used across all samples in a batch, which leads to severe performance collapse, highlighting the importance of input-adaptive routing. Removing layer-wise router enforces fixed expert assignment across all MoE-LayerNorm layers, limiting the model’s flexibility and hurting performance. Eliminating the entire MoE-LayerNorm module results in unstable and ineffective adaptation, confirming its necessity.



\begin{figure*}[ht]
    \centering
    \includegraphics{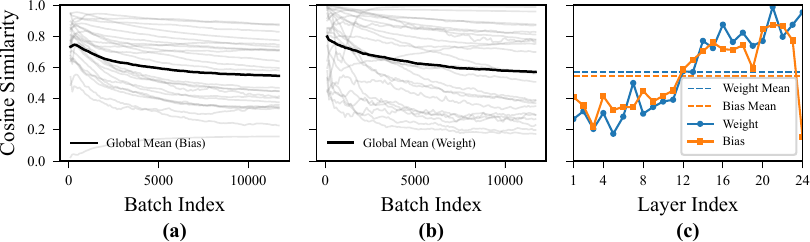}
  \caption{    
  Evolution and final-state statistics of expert parameter similarity across MoE-LayerNorms adapted under classical mixed distribution shifts. (a–b) track the average pairwise cosine similarity of expert bias and weight parameters during adaptation, with each gray curve representing one MoE-LayerNorm and the bold black line denoting the global mean across layers. (c) reports the final average similarity for each layer at the end of adaptation phase.}
  \label{fig:moe_similarity}
\end{figure*}

\subsectionwithouttoc{Analysis of Expert Parameter Diversity}

Fig.~\ref{fig:moe_similarity} shows that, during adaptation, experts' parameters within the same MoE-LayerNorm diverge over time. This effect is particularly pronounced in shallow layers, reflected by the decreasing cosine similarity in Fig.~\ref{fig:moe_similarity}(a-b). These results suggest that experts learn distinct parameter update trajectories, rather than converging to similar representations.

Importantly, such diversity is not pre-imposed but emerges naturally from our design: experts are structurally decoupled and trained with separate gradient signals via the routing mechanism. The final-state statistics in Fig.~\ref{fig:moe_similarity}(c) confirm this decoupling, with many layers exhibiting low similarity among their experts.

This parameter-level diversity allows MoETTA to internalize multiple adaptation modes within a single model, enabling it to better accommodate heterogeneous input batches without explicit domain labels. A  layer-wise visualization of expert similarity is included in App.~\ref{app:cosine_similarity_visualization}.

\begin{figure}[tbp]
  \centering
    \includegraphics{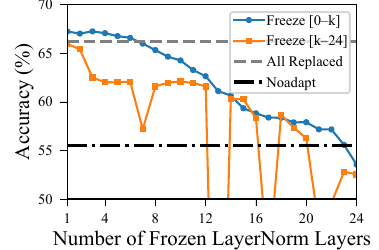}
    \caption{
Effect of partially replacing LayerNorm with MoE-LayerNorm on accuracy.
}
    \label{fig:partial_replacement}
\end{figure}

\subsectionwithouttoc{Hyper-Parameter Sensitivity}
\label{subsec:hyper}
In this subsection, all experiments are conducted under the classical mixed distribution shifts.

\paragraph{Effect of Replacement Strategy.}

As shown in Fig.~\ref{fig:partial_replacement}, we compare two partial replacement strategies for MoE-LayerNorm: (1) freezing shallow LayerNorm layers indexed $0$–$k$, and (2) freezing deeper layers indexed $k$–$24$. 
Freezing the shallow layers ($k = 0,\ldots,5$) consistently outperforms full replacement. This is consistent with prior findings that shallow ViT layers primarily encode low-level, domain-invariant features such as color~\citep{ViT_early_layer}, which benefit from being preserved. In contrast, allowing deeper layers—responsible for high-level semantics—to adapt leads to more effective test-time adaptation.
Freezing middle layers, which are known to be critical for fine-tuning~\citep{ViT_middle_layer}, results in a sharp performance drop. Similarly, freezing only the deeper layers hinders the model’s ability to adjust to distribution shifts, often leading to degraded performance or collapse.

\begin{figure}[tbp]
    \centering
    \includegraphics{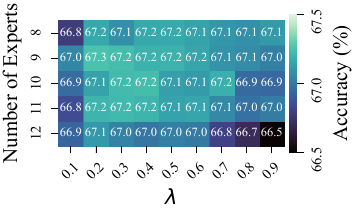}
    \caption{Grid search results over the number of experts and the trade-off coefficient $\lambda$ in Eq.~\eqref{eq:lb_alpha}.}
    \label{fig:sensitivity}
\end{figure}

\paragraph{Effect of the Number of Experts and $\lambda$ in Eq.~\eqref{eq:lb_alpha}.}

As shown in Fig.~\ref{fig:sensitivity}, MoETTA performs best when the trade-off coefficient $\lambda$ is small, suggesting that excessive focus on load balancing can impede effective adaptation. This underscores the value of letting the router flexibly assign experts based on input characteristics rather than enforcing uniform expert usage.
The number of experts also affects performance. While MoETTA is robust across a range of choices, using 9 experts achieves the best accuracy under classical mixed distribution shifts.

\sectionwithouttoc{Conclusion}
\label{conclusion_and_limitation}

We study TTA under mixed distribution shifts, where a single update direction across diverse samples often leads to suboptimal adaptation. For more realistic evaluation, we introduce two benchmarks, potpourri and potpourri+, combining multiple shift types within a single test stream.

To overcome the limitations of conventional TTA, we propose MoETTA, a lightweight framework that reparameterizes LayerNorm into a mixture of decoupled expert branches. Allowing experts to follow distinct adaptation trajectories lets MoETTA capture multiple modes of test-time behavior. Empirically, this expert diversity improves robustness and stability under complex mixed distribution shifts, consistently outperforming prior TTA methods.

\section*{Acknowledgments}
This study was supported by the National Key Research and Development Program of China (Grant No. 2023YFF0905502), the National Natural Science Foundation of China (Grant Nos. 92467204 and 62472249), the Shenzhen Science and Technology Program (Grant Nos. JCYJ20220818101014030, KJZD20240903102300001, and JCYJ20250604145014018), and the Natural Science Foundation for Top Talents of Shenzhen Technology University (Grant No. GDRC202413).
\bibliography{bib/aaai2026}

\begin{thebibliography}{44}
\providecommand{\natexlab}[1]{#1}

\bibitem[{Ba, Kiros, and Hinton(2016)}]{layernorm}
Ba, J.~L.; Kiros, J.~R.; and Hinton, G.~E. 2016.
\newblock Layer Normalization.
\newblock \emph{arXiv preprint arXiv:1607.06450}.

\bibitem[{Ben-David et~al.(2010)Ben-David, Blitzer, Crammer, Kulesza, Pereira, and Vaughan}]{domain_theory}
Ben-David, S.; Blitzer, J.; Crammer, K.; Kulesza, A.; Pereira, F.; and Vaughan, J. 2010.
\newblock A theory of learning from different domains.
\newblock \emph{Machine Learning}, 79: 151--175.

\bibitem[{Cang, Chen, and Huang(2024)}]{joint_batch}
Cang, Y.; Chen, M.; and Huang, K. 2024.
\newblock Joint Batching and Scheduling for High-Throughput Multiuser Edge AI With Asynchronous Task Arrivals.
\newblock \emph{IEEE Transactions on Wireless Communications}, 23(10): 13782--13795.

\bibitem[{Chen et~al.(2024{\natexlab{a}})Chen, Niu, Chen, Zhang, Li, Li, and Tan}]{CoLa}
Chen, G.; Niu, S.; Chen, D.; Zhang, S.; Li, C.; Li, Y.; and Tan, M. 2024{\natexlab{a}}.
\newblock Cross-Device Collaborative Test-Time Adaptation.
\newblock In \emph{The Thirty-eighth Annual Conference on Neural Information Processing Systems}.

\bibitem[{Chen et~al.(2024{\natexlab{b}})Chen, Niu, Xu, Song, Wang, and Tan}]{CEMA}
Chen, Y.; Niu, S.; Xu, S.; Song, H.; Wang, Y.; and Tan, M. 2024{\natexlab{b}}.
\newblock Towards Robust and Efficient Cloud-Edge Elastic Model Adaptation via Selective Entropy Distillation.
\newblock In \emph{International Conference on Learning Representations}.

\bibitem[{Dai et~al.(2024)Dai, Deng, Zhao, Xu, Gao, Chen, Li, Zeng, Yu, Wu, Xie, Li, Huang, Luo, Ruan, Sui, and Liang}]{deepseekmoe}
Dai, D.; Deng, C.; Zhao, C.; Xu, R.; Gao, H.; Chen, D.; Li, J.; Zeng, W.; Yu, X.; Wu, Y.; Xie, Z.; Li, Y.; Huang, P.; Luo, F.; Ruan, C.; Sui, Z.; and Liang, W. 2024.
\newblock {D}eep{S}eek{M}o{E}: Towards Ultimate Expert Specialization in Mixture-of-Experts Language Models.
\newblock In Ku, L.-W.; Martins, A.; and Srikumar, V., eds., \emph{Proceedings of the 62nd Annual Meeting of the Association for Computational Linguistics (Volume 1: Long Papers)}, 1280--1297. Bangkok, Thailand: Association for Computational Linguistics.

\bibitem[{DeepSeek-AI(2024)}]{DeepSeekV3}
DeepSeek-AI. 2024.
\newblock DeepSeek-V3 Technical Report.
\newblock arXiv:2412.19437.

\bibitem[{Deng et~al.(2009)Deng, Dong, Socher, Li, Li, and Fei-Fei}]{imagenet}
Deng, J.; Dong, W.; Socher, R.; Li, L.-J.; Li, K.; and Fei-Fei, L. 2009.
\newblock Imagenet: {A} Large-Scale Hierarchical Image Database.
\newblock In \emph{CVPR}, 248--255.

\bibitem[{Deng et~al.(2025)Deng, Niu, Zhang, Chen, Zeng, Chen, and Hu}]{MGTTA}
Deng, Q.; Niu, S.; Zhang, R.; Chen, Y.; Zeng, R.; Chen, J.; and Hu, X. 2025.
\newblock Learning to Generate Gradients for Test-Time Adaptation via Test-Time Training Layers.
\newblock In \emph{Proceedings of the AAAI Conference on Artificial Intelligence}.

\bibitem[{Dorszewski et~al.(2025)Dorszewski, T{\v{e}}tkov{\'a}, Jenssen, Hansen, and Wickstr{\o}m}]{ViT_early_layer}
Dorszewski, T.; T{\v{e}}tkov{\'a}, L.; Jenssen, R.; Hansen, L.~K.; and Wickstr{\o}m, K.~K. 2025.
\newblock From colors to classes: Emergence of concepts in vision transformers.
\newblock In \emph{World Conference on Explainable Artificial Intelligence}, 28--47. Springer.

\bibitem[{Dosovitskiy et~al.(2021)Dosovitskiy, Beyer, Kolesnikov, Weissenborn, Zhai, Unterthiner, Dehghani, Minderer, Heigold, Gelly, Uszkoreit, and Houlsby}]{ViT}
Dosovitskiy, A.; Beyer, L.; Kolesnikov, A.; Weissenborn, D.; Zhai, X.; Unterthiner, T.; Dehghani, M.; Minderer, M.; Heigold, G.; Gelly, S.; Uszkoreit, J.; and Houlsby, N. 2021.
\newblock An Image is Worth 16x16 Words: Transformers for Image Recognition at Scale.
\newblock \emph{ICLR}.

\bibitem[{Eigen, Ranzato, and Sutskever(2014)}]{eigen2014learningfactoredrepresentationsdeep}
Eigen, D.; Ranzato, M.; and Sutskever, I. 2014.
\newblock Learning Factored Representations in a Deep Mixture of Experts.
\newblock arXiv:1312.4314.

\bibitem[{Fedus, Zoph, and Shazeer(2022)}]{switch_transformer}
Fedus, W.; Zoph, B.; and Shazeer, N. 2022.
\newblock Switch transformers: scaling to trillion parameter models with simple and efficient sparsity.
\newblock \emph{J. Mach. Learn. Res.}, 23(1).

\bibitem[{Foret et~al.(2021)Foret, Kleiner, Mobahi, and Neyshabur}]{SAM}
Foret, P.; Kleiner, A.; Mobahi, H.; and Neyshabur, B. 2021.
\newblock Sharpness-aware Minimization for Efficiently Improving Generalization.
\newblock In \emph{International Conference on Learning Representations}.

\bibitem[{Gidaris, Singh, and Komodakis(2018)}]{gidaris2018unsupervised}
Gidaris, S.; Singh, P.; and Komodakis, N. 2018.
\newblock Unsupervised Representation Learning by Predicting Image Rotations.
\newblock In \emph{International Conference on Learning Representations}.

\bibitem[{Glorot and Bengio(2010)}]{xavier}
Glorot, X.; and Bengio, Y. 2010.
\newblock Understanding the difficulty of training deep feedforward neural networks.
\newblock In Teh, Y.~W.; and Titterington, M., eds., \emph{Proceedings of the Thirteenth International Conference on Artificial Intelligence and Statistics}, volume~9 of \emph{Proceedings of Machine Learning Research}, 249--256. Chia Laguna Resort, Sardinia, Italy: PMLR.

\bibitem[{Grandvalet and Bengio(2004)}]{entropy}
Grandvalet, Y.; and Bengio, Y. 2004.
\newblock Semi-supervised Learning by Entropy Minimization.
\newblock In Saul, L.; Weiss, Y.; and Bottou, L., eds., \emph{Advances in Neural Information Processing Systems}, volume~17. MIT Press.

\bibitem[{Hendrycks et~al.(2021{\natexlab{a}})Hendrycks, Basart, Mu, Kadavath, Wang, Dorundo, Desai, Zhu, Parajuli, Guo et~al.}]{imagenet-r}
Hendrycks, D.; Basart, S.; Mu, N.; Kadavath, S.; Wang, F.; Dorundo, E.; Desai, R.; Zhu, T.; Parajuli, S.; Guo, M.; et~al. 2021{\natexlab{a}}.
\newblock The many faces of robustness: A critical analysis of out-of-distribution generalization.
\newblock In \emph{Proceedings of the IEEE/CVF International Conference on Computer Vision}, 8340--8349.

\bibitem[{Hendrycks and Dietterich(2019)}]{imagenet-c}
Hendrycks, D.; and Dietterich, T. 2019.
\newblock Benchmarking Neural Network Robustness to Common Corruptions and Perturbations.
\newblock In \emph{International Conference on Learning Representations}.

\bibitem[{Hendrycks et~al.(2021{\natexlab{b}})Hendrycks, Zhao, Basart, Steinhardt, and Song}]{imagenet-a}
Hendrycks, D.; Zhao, K.; Basart, S.; Steinhardt, J.; and Song, D. 2021{\natexlab{b}}.
\newblock Natural Adversarial Examples.
\newblock \emph{CVPR}.

\bibitem[{Hu et~al.(2022)Hu, Shen, Wallis, Allen-Zhu, Li, Wang, Wang, Chen et~al.}]{hu2022lora}
Hu, E.~J.; Shen, Y.; Wallis, P.; Allen-Zhu, Z.; Li, Y.; Wang, S.; Wang, L.; Chen, W.; et~al. 2022.
\newblock Lora: Low-rank adaptation of large language models.
\newblock \emph{ICLR}, 1(2): 3.

\bibitem[{Jacobs et~al.(1991)Jacobs, Jordan, Nowlan, and Hinton}]{jacobs1991adaptive}
Jacobs, R.~A.; Jordan, M.~I.; Nowlan, S.~J.; and Hinton, G.~E. 1991.
\newblock Adaptive mixtures of local experts.
\newblock \emph{Neural computation}, 3(1): 79--87.

\bibitem[{Jiang et~al.(2024)Jiang, Sablayrolles, Roux, Mensch, Savary, Bamford, Chaplot, de~las Casas, Hanna, Bressand, Lengyel, Bour, Lample, Lavaud, Saulnier, Lachaux, Stock, Subramanian, Yang, Antoniak, Scao, Gervet, Lavril, Wang, Lacroix, and Sayed}]{mixtral}
Jiang, A.~Q.; Sablayrolles, A.; Roux, A.; Mensch, A.; Savary, B.; Bamford, C.; Chaplot, D.~S.; de~las Casas, D.; Hanna, E.~B.; Bressand, F.; Lengyel, G.; Bour, G.; Lample, G.; Lavaud, L.~R.; Saulnier, L.; Lachaux, M.-A.; Stock, P.; Subramanian, S.; Yang, S.; Antoniak, S.; Scao, T.~L.; Gervet, T.; Lavril, T.; Wang, T.; Lacroix, T.; and Sayed, W.~E. 2024.
\newblock Mixtral of Experts.
\newblock arXiv:2401.04088.

\bibitem[{Jordan and Jacobs(1994)}]{jordan1994hierarchical}
Jordan, M.~I.; and Jacobs, R.~A. 1994.
\newblock Hierarchical mixtures of experts and the EM algorithm.
\newblock \emph{Neural computation}, 6(2): 181--214.

\bibitem[{Lee, Yoon, and Hwang(2024)}]{BECoTTA}
Lee, D.; Yoon, J.; and Hwang, S.~J. 2024.
\newblock BECoTTA: Input-dependent Online Blending of Experts for Continual Test-time Adaptation.
\newblock In \emph{International Conference on Machine Learning}.

\bibitem[{Lee et~al.(2024)Lee, Jung, Lee, Park, Shin, Hwang, and Yoon}]{DeYO}
Lee, J.; Jung, D.; Lee, S.; Park, J.; Shin, J.; Hwang, U.; and Yoon, S. 2024.
\newblock Entropy is not Enough for Test-Time Adaptation: From the Perspective of Disentangled Factors.
\newblock In \emph{The Twelfth International Conference on Learning Representations}.

\bibitem[{Liu et~al.(2021)Liu, Kothari, van Delft, Bellot-Gurlet, Mordan, and Alahi}]{TTT1}
Liu, Y.; Kothari, P.; van Delft, B.; Bellot-Gurlet, B.; Mordan, T.; and Alahi, A. 2021.
\newblock TTT++: When Does Self-Supervised Test-Time Training Fail or Thrive?
\newblock In Ranzato, M.; Beygelzimer, A.; Dauphin, Y.; Liang, P.; and Vaughan, J.~W., eds., \emph{Advances in Neural Information Processing Systems}, volume~34, 21808--21820. Curran Associates, Inc.

\bibitem[{Liu et~al.(2022)Liu, Mao, Wu, Feichtenhofer, Darrell, and Xie}]{liu2022convnet}
Liu, Z.; Mao, H.; Wu, C.-Y.; Feichtenhofer, C.; Darrell, T.; and Xie, S. 2022.
\newblock A convnet for the 2020s.
\newblock In \emph{Proceedings of the IEEE/CVF conference on computer vision and pattern recognition}, 11976--11986.

\bibitem[{Niu et~al.(2022)Niu, Wu, Zhang, Chen, Zheng, Zhao, and Tan}]{EATA}
Niu, S.; Wu, J.; Zhang, Y.; Chen, Y.; Zheng, S.; Zhao, P.; and Tan, M. 2022.
\newblock Efficient Test-Time Model Adaptation without Forgetting.
\newblock In \emph{The International Conference on Machine Learning}.

\bibitem[{Niu et~al.(2023)Niu, Wu, Zhang, Wen, Chen, Zhao, and Tan}]{SAR}
Niu, S.; Wu, J.; Zhang, Y.; Wen, Z.; Chen, Y.; Zhao, P.; and Tan, M. 2023.
\newblock Towards Stable Test-Time Adaptation in Dynamic Wild World.
\newblock In \emph{International Conference on Learning Representations}.

\bibitem[{Paszke et~al.(2019)Paszke, Gross, Massa, Lerer, Bradbury, Chanan, Killeen, Lin, Gimelshein, Antiga, Desmaison, K\"{o}pf, Yang, DeVito, Raison, Tejani, Chilamkurthy, Steiner, Fang, Bai, and Chintala}]{Pytorch}
Paszke, A.; Gross, S.; Massa, F.; Lerer, A.; Bradbury, J.; Chanan, G.; Killeen, T.; Lin, Z.; Gimelshein, N.; Antiga, L.; Desmaison, A.; K\"{o}pf, A.; Yang, E.; DeVito, Z.; Raison, M.; Tejani, A.; Chilamkurthy, S.; Steiner, B.; Fang, L.; Bai, J.; and Chintala, S. 2019.
\newblock \emph{PyTorch: an imperative style, high-performance deep learning library}, 12.
\newblock Red Hook, NY, USA: Curran Associates Inc.

\bibitem[{Shazeer et~al.(2017{\natexlab{a}})Shazeer, Mirhoseini, Maziarz, Davis, Le, Hinton, and Dean}]{shazeer2017outrageously}
Shazeer, N.; Mirhoseini, A.; Maziarz, K.; Davis, A.; Le, Q.; Hinton, G.; and Dean, J. 2017{\natexlab{a}}.
\newblock Outrageously large neural networks: The sparsely-gated mixture-of-experts layer.
\newblock \emph{arXiv preprint arXiv:1701.06538}.

\bibitem[{Shazeer et~al.(2017{\natexlab{b}})Shazeer, Mirhoseini, Maziarz, Davis, Le, Hinton, and Dean}]{MoE}
Shazeer, N.; Mirhoseini, A.; Maziarz, K.; Davis, A.; Le, Q.; Hinton, G.; and Dean, J. 2017{\natexlab{b}}.
\newblock Outrageously Large Neural Networks: The Sparsely-Gated Mixture-of-Experts Layer.
\newblock In \emph{International Conference on Learning Representations}.

\bibitem[{Su et~al.(2024)Su, Guo, Yao, Yang, Wang, and Huang}]{use_mix_scenario_1}
Su, Z.; Guo, J.; Yao, K.; Yang, X.; Wang, Q.; and Huang, K. 2024.
\newblock Unraveling Batch Normalization for Realistic Test-Time Adaptation.
\newblock In Wooldridge, M.~J.; Dy, J.~G.; and Natarajan, S., eds., \emph{Thirty-Eighth {AAAI} Conference on Artificial Intelligence, {AAAI} 2024, Thirty-Sixth Conference on Innovative Applications of Artificial Intelligence, {IAAI} 2024, Fourteenth Symposium on Educational Advances in Artificial Intelligence, {EAAI} 2014, February 20-27, 2024, Vancouver, Canada}, 15136--15144. {AAAI} Press.

\bibitem[{Tan et~al.(2025)Tan, Chen, Wu, Zhang, Chen, Zhao, and Niu}]{EATA-C}
Tan, M.; Chen, G.; Wu, J.; Zhang, Y.; Chen, Y.; Zhao, P.; and Niu, S. 2025.
\newblock Uncertainty-Calibrated Test-Time Model Adaptation without Forgetting.
\newblock \emph{IEEE Transactions on Pattern Analysis and Machine Intelligence}, 1--14.

\bibitem[{Valeriani et~al.(2023)Valeriani, Doimo, Cuturello, Laio, Ansuini, and Cazzaniga}]{ViT_middle_layer}
Valeriani, L.; Doimo, D.; Cuturello, F.; Laio, A.; Ansuini, A.; and Cazzaniga, A. 2023.
\newblock The geometry of hidden representations of large transformer models.
\newblock In \emph{Proceedings of the 37th International Conference on Neural Information Processing Systems}, NIPS '23. Red Hook, NY, USA: Curran Associates Inc.

\bibitem[{van~der Maaten and Hinton(2008)}]{tsne}
van~der Maaten, L.; and Hinton, G.~E. 2008.
\newblock Visualizing Data using t-SNE.
\newblock \emph{Journal of Machine Learning Research}, 9: 2579--2605.

\bibitem[{Vaswani et~al.(2017)Vaswani, Shazeer, Parmar, Uszkoreit, Jones, Gomez, Kaiser, and Polosukhin}]{attention}
Vaswani, A.; Shazeer, N.; Parmar, N.; Uszkoreit, J.; Jones, L.; Gomez, A.~N.; Kaiser, L.; and Polosukhin, I. 2017.
\newblock Attention is all you need.
\newblock In \emph{Proceedings of the 31st International Conference on Neural Information Processing Systems}, NIPS'17, 6000–6010. Red Hook, NY, USA: Curran Associates Inc.
\newblock ISBN 9781510860964.

\bibitem[{Wang et~al.(2021)Wang, Shelhamer, Liu, Olshausen, and Darrell}]{Tent}
Wang, D.; Shelhamer, E.; Liu, S.; Olshausen, B.; and Darrell, T. 2021.
\newblock Tent: Fully Test-Time Adaptation by Entropy Minimization.
\newblock In \emph{International Conference on Learning Representations}.

\bibitem[{Wang et~al.(2019)Wang, Ge, Lipton, and Xing}]{imagenet-sketch}
Wang, H.; Ge, S.; Lipton, Z.; and Xing, E.~P. 2019.
\newblock Learning robust global representations by penalizing local predictive power.
\newblock In \emph{Advances in Neural Information Processing Systems}, 10506--10518.

\bibitem[{Wang et~al.(2022)Wang, Fink, Van~Gool, and Dai}]{COTTA}
Wang, Q.; Fink, O.; Van~Gool, L.; and Dai, D. 2022.
\newblock Continual Test-Time Domain Adaptation.
\newblock In \emph{Proceedings of Conference on Computer Vision and Pattern Recognition}.

\bibitem[{Wang et~al.(2024)Wang, Luo, Zheng, Chen, Wang, and Huang}]{wang2024search}
Wang, Z.; Luo, Y.; Zheng, L.; Chen, Z.; Wang, S.; and Huang, Z. 2024.
\newblock In Search of Lost Online Test-Time Adaptation: A Survey.
\newblock \emph{International Journal of Computer Vision}, 133: 1106--1139.

\bibitem[{Wightman(2019)}]{timm}
Wightman, R. 2019.
\newblock PyTorch Image Models.
\newblock \url{https://github.com/rwightman/pytorch-image-models}.

\bibitem[{Wiles et~al.(2022)Wiles, Gowal, Stimberg, Rebuffi, Ktena, Dvijotham, and Cemgil}]{spurious_distribution_shift}
Wiles, O.; Gowal, S.; Stimberg, F.; Rebuffi, S.-A.; Ktena, I.; Dvijotham, K.~D.; and Cemgil, A.~T. 2022.
\newblock A Fine-Grained Analysis on Distribution Shift.
\newblock In \emph{International Conference on Learning Representations}.

\end{thebibliography}
\clearpage
\newpage
\onecolumn
\setcounter{secnumdepth}{2}

\renewcommand{\thesection}{\Alph{section}}
\setcounter{section}{0}  

\renewcommand{\theequation}{\arabic{equation}}
\renewcommand{\thefigure}{\arabic{figure}}
\renewcommand{\thetable}{\arabic{table}}

\section*{Appendix}

\setcounter{equation}{0}
\setcounter{figure}{0}
\setcounter{table}{0}
In this appendix, we provide additional experimental details and results. The appendix is organized into the following sections:
\begin{itemize}
\item Section~\ref{app:pseudo_code} presents the pseudo code for the forward pass of MoE-LayerNorm, as well as the overall adaptation process of our method.
\item Section~\ref{app:cosine_similarity_proof} provides a proof for the convergence of the expectation of the cosine similarity of the experts.
\item Section~\ref{app:load_balancing} presents a proof for the lower bound of the load balancing loss.
\item Section~\ref{app:dataset} provides information about the datasets used in this paper.
\item Section~\ref{app:implementaion_details} outlines the implementation details.
\item Section~\ref{app:compute_resource} describes the computing resources used throughout the experiments.
\item Section~\ref{app:cosine_similarity_visualization} presents the visualization of the cosine similarity of the experts within each MoE-LayerNorm.
\item Section~\ref{app:grid_search} reports the results of the grid search conducted for the baselines.
\item Section~\ref{app:scalability} reports the results using ViT-L/16 to show the scalability of MoETTA.
\item Section~\ref{app:broader_impacts} discusses the potential broader impacts of our research.
\end{itemize}
\clearpage
\newpage
\section{Pseudocode for Adaptation Process and MoE-LayerNorm}
\label{app:pseudo_code}
We provide the pseudocode for the overall test-time adaptation process, along with a PyTorch~\citep{Pytorch} style implementation of the forward pass for MoE-LayerNorm. To enhance generality, the MoE-LayerNorm pseudocode supports activating multiple experts per sample, rather than just one.
\vfill
\begin{algorithm*}
    \KwIn{Test samples $\mathcal{D}_{test}=\{\{\mbf x^{(i)}_j\}_{j=1}^B\}_{i=1}^{N}$, the pre-trained model $f(\cdot;\theta)$, number of classes $N$, hyper-parameters $\lambda$, $E_0$.}
    \KwOut{The predictions $\{\{\hat{y}_j^{(i)}\}_{j=1}^B\}_{i=1}^N$}
    Replace LayerNorm in $f(\cdot; \theta)$ with MoELayerNorm.\;
    $entropys \gets$ [\ ]\;
    \For{$i\text{-th batch }\{\mbf x^{(i)}_j\}_{j=1}^B$ \InKw $\mathcal{D}_{test}$}
    {
        Calculate posterior probability : $\{\{\hat{p_k}\}_{k=1}^N\}_{j=1}^B\gets f(\{\mbf x^{(i)}_j\}_{j=1}^B; \theta)$\;
        Get predictions for $i\text{-th batch}$ : $\{\hat{y}_j^{(i)}\}_{j=1}^B\gets\arg\max_k\{\{\hat{p_k}\}_{k=1}^N\}_{j=1}^B$\;
        Calculate entropy : $\{e_j\}_{j=1}^B \gets entropy(\{\{\hat{p_k}\}_{k=1}^N\}_{j=1}^B)$\;
        $mean\_entropy \gets \sum_{j=1}^Be_j/B$\;
        \eIf{i==1}{
            $entropys.append(mean\_entropy)$\;
            $threshold\gets mean\_entropy$\;
            $\alpha\gets \lambda\cdot mean\_entropy$\;
        }
        {
            $old\_avg \gets mean(entropys)$\;
            $entropys.append(mean\_entropy)$\;
            $new\_avg \gets mean(entropys)$\;
            $threshold\gets threshold\cdot new\_avg/old\_avg$\;
            $\alpha\gets \alpha\cdot new\_avg/old\_avg$\;
        }
        $loss \gets \sum_{i=1}^B[\mathbb{I}_{\{e_i<threshold\}}\exp\left[E_0-e_i.detach()\right]\cdot e_i]/\sum_{i=1}^B\mathbb{I}_{\{e_i<threshold\}}$\;
        $load\_balancing\_loss \gets collect\_load\_balancing\_loss(f(\cdot; \theta))$\;
        $(loss+\alpha\cdot load\_balancing\_loss).backward()$\;
    }
    \caption{Adaptation process}
\end{algorithm*}
\vfill
\begin{algorithm*}
    \KwIn{Input $X \in \mathbb{R}^{B \times T \times D}$, expert weights $W_e \in \mathbb{R}^{E \times D}$, biases $b_e \in \mathbb{R}^{E \times D}$, shared weight $W \in \mathbb{R}^D$, bias $b \in \mathbb{R}^D$, router $g: \mathbb{R}^D \rightarrow \mathbb{R}^E$, number of activated experts $k$}
    \KwOut{Layer-normalized tensor $Y \in \mathbb{R}^{B \times T \times D}$}

    $Z \gets \mathrm{mean}(X, \text{dim}=1)$ \tcp*{Shape: $B \times D$}
    $P \gets \mathrm{softmax}(g(Z))$ \tcp*{Router probability, shape: $B \times E$}
    $\texttt{topk\_idx} \gets \mathrm{TopK}(P, k)$ \tcp*{Shape: $B \times k$}
    $\texttt{prob} \gets \texttt{gather}(P, \texttt{topk\_idx})$ \tcp*{Shape: $B \times k$}
    $\texttt{importance} \gets \mathrm{mean}(P, \text{dim}=0)$ \tcp*{Expert mean routing prob}
    $\texttt{count} \gets \texttt{bincount}(\texttt{topk\_idx})$ \tcp*{Expert assignment count}
    $\texttt{load} \gets \texttt{count} / \texttt{sum}(\texttt{count})$ \tcp*{Normalized expert usage}
    $\texttt{aux\_loss} \gets E \cdot \texttt{sum}(\texttt{importance} \cdot \texttt{load})$ \tcp*{To be collected later}
    $c \gets \texttt{prob} / \texttt{prob.detach()}$ \tcp*{Forward as 1, backward to router}
    $W_{\text{sel}} \gets \texttt{gather}(W_e, \texttt{topk\_idx})$ \tcp*{Shape: $B \times k \times D$}
    $b_{\text{sel}} \gets \texttt{gather}(b_e, \texttt{topk\_idx})$ \tcp*{Shape: $B \times k \times D$}
    $c \gets \texttt{unsqueeze}(c, -1)$ \tcp*{Broadcast: $B \times k \times 1$}
    $W_{\text{fused}} \gets \mathrm{sum}(W_{\text{sel}} \cdot c, \text{dim}=1) + W$ \tcp*{Shape: $B \times D$}
    $b_{\text{fused}} \gets \mathrm{sum}(b_{\text{sel}} \cdot c, \text{dim}=1) + b$ \tcp*{Shape: $B \times D$}
    $\mu \gets \mathrm{mean}(X, \text{dim}=-1, \text{keepdim}=True)$ \tcp*{Shape: $B \times T \times 1$}
    $\sigma^2 \gets \mathrm{var}(X, \text{dim}=-1, \text{keepdim}=True)$\tcp*{Shape: $B \times T \times 1$}
    $X_{\text{norm}} \gets (X - \mu) / \sqrt{\sigma^2 + \epsilon}$\tcp*{Shape: $B \times T \times D$}
    $Y \gets X_{\text{norm}} \cdot W_{\text{fused}}[:, \texttt{None}, :] + b_{\text{fused}}[:, \texttt{None}, :]$\tcp*{Shape: $B \times T \times D$}
    \caption{MoE-LayerNorm}
\end{algorithm*}
\clearpage
\newpage
\section{Theoretical Result: Cosine Similarity Expectation}
\label{app:cosine_similarity_proof}

We present a theoretical result on the expected cosine similarity between two independent high-dimensional Gaussian vectors, shifted by a common reference. All assumptions are explicitly stated, and a complete proof is provided below.

\begin{proposition}\label{prop:cos_expect_en}
Let $\theta_1,\theta_2,\theta\in\mathbb{R}^d$ be independent random vectors sampled as
\[
\theta_1,\theta_2,\theta\;\stackrel{\text{i.i.d.}}{\sim}\mathcal N(\mathbf 0,\sigma^2 I_d),
\]
and define
\[
\mathbf u:=\theta_1-\theta,\qquad 
\mathbf v:=\theta_2-\theta .
\]
Then, the expected cosine similarity satisfies
\[
\mathbb E\!\Bigl[\cos\!\bigl\langle\mathbf u,\mathbf v\bigr\rangle\Bigr]
=\frac12+\mathcal O\!\bigl(d^{-1}\bigr),
\qquad \text{i.e.},\;
\lim_{d\to\infty}\mathbb E\bigl[\cos\langle\mathbf u,\mathbf v\rangle\bigr]=\tfrac12 .
\]
Moreover, this expectation is independent of $\sigma$.
\end{proposition}

\begin{proof}
\textbf{Covariance structure.}
For each coordinate $i=1,\dots,d$, we write:
\[
u_i=\theta_{1,i}-\theta_i,\qquad 
v_i=\theta_{2,i}-\theta_i.
\]
By independence:
\[
\operatorname{Var}(u_i)=\operatorname{Var}(v_i)=2\sigma^2,
\quad
\operatorname{Cov}(u_i,v_i)=\sigma^2 .
\]
Thus,
\[
(u_i,v_i)\sim\mathcal N\!\left(\mathbf 0,
\sigma^2
\begin{pmatrix}
 2 & 1\\[2pt]
 1 & 2
\end{pmatrix}\right),
\]
yielding a per-coordinate correlation coefficient $\rho=\tfrac12$.

\textbf{High-dimensional limit.}
Normalize:
\[
\mathbf u'=\frac{\mathbf u}{\sqrt{2}\sigma},
\quad 
\mathbf v'=\frac{\mathbf v}{\sqrt{2}\sigma},
\]
so that all coordinates have unit variance and correlation $\rho=\tfrac12$. Let
\[
s_d:=\frac{\mathbf u'^{\!\top}\mathbf v'}%
           {\lVert\mathbf u'\rVert\,\lVert\mathbf v'\rVert}
     =\cos\langle\mathbf u',\mathbf v'\rangle
     =\cos\langle\mathbf u,\mathbf v\rangle.
\]
By the strong law of large numbers:
\[
\frac{\mathbf u'^{\!\top}\mathbf v'}{d}
      =\frac1d\sum_{i=1}^{d}u_i'v_i'
      \xrightarrow{\text{a.s.}}\rho,
\qquad
\frac{\lVert\mathbf u'\rVert^{2}}{d}
      \xrightarrow{\text{a.s.}}1,
\qquad
\frac{\lVert\mathbf v'\rVert^{2}}{d}
      \xrightarrow{\text{a.s.}}1,
\]
hence
\[
s_d\;\xrightarrow{\text{a.s.}}\;\rho=\tfrac12.
\]

\textbf{Expectation.}
Since $s_d \in [-1,1]$, by the dominated convergence theorem:
\[
\mathbb E[s_d]\;\longrightarrow\;\tfrac12.
\]
Furthermore, a refined Berry–Esseen-type estimate gives
\(
\mathbb E[s_d]=\tfrac12+\mathcal O(d^{-1}).
\)

\textbf{Independence of $\sigma$.}
We observe that in
\[
\cos\langle\mathbf u,\mathbf v\rangle
=\frac{\mathbf u^{\!\top}\mathbf v}
       {\lVert\mathbf u\rVert\,\lVert\mathbf v\rVert},
\]
both numerator and denominator scale linearly with $\sigma$, so the expectation is invariant to the choice of $\sigma$.
\end{proof}

\clearpage
\newpage
\section{Theoretical Result: the Lower Bound of the Load Balancing Loss}
\label{app:load_balancing}

We provide a formal proof of the lower bound of the load balancing loss under Top-1 routing, following the definition introduced by \citet{switch_transformer}.

\begin{proposition}\label{prop:lb_lower_bound}
With the Top-1 routing rule~\citep{switch_transformer}, the auxiliary load balancing loss
\[
\mathcal{L}_{\text{load balancing}}
   \;=\; N\times\sum_{i=1}^{N} f_i\times P_i
\]
is lower bounded by \(1\) for \emph{every} mini-batch \(\mathcal B\); equality holds if and only if every sample's routing probabilities are uniform:
\[
p_1(\mathbf x) = \dots = p_N(\mathbf x) = \tfrac{1}{N}.
\]
\end{proposition}

\begin{proof}
Let \(\mathcal B = \{\mathbf x_1, \dots, \mathbf x_{|\mathcal B|}\}\) be a mini-batch. Denote the selected expert for each input as
\begin{equation}
i(\mathbf x) = \arg\max_{k} p_k(\mathbf x),
\qquad k \in \{1, \dots, N\}.
\label{eq:argmax_routing}
\end{equation}
By construction, the Top-1 router guarantees that for every \(\mathbf x\),
\begin{equation}
p_{i(\mathbf x)}(\mathbf x) \ge \frac{1}{N},
\label{eq:min_max_prob}
\end{equation}
since the maximum of \(N\) non-negative values summing to 1 must be at least \(1/N\).

Define the following quantity:
\begin{equation}
Z := \frac{1}{|\mathcal B|} \sum_{\mathbf x \in \mathcal B} p_{i(\mathbf x)}(\mathbf x).
\label{eq:z_definition}
\end{equation}
Using inequality~\eqref{eq:min_max_prob}, we obtain:
\begin{equation}
Z \ge \frac{1}{N}.
\label{eq:z_bound}
\end{equation}

Next, rewrite \(Z\) in terms of \(f_i\) and \(P_i\), where:
\[
f_i := \frac{1}{|\mathcal B|} \sum_{\mathbf x \in \mathcal B} \mathbf{1}_{i(\mathbf x) = i},
\qquad
P_i := \frac{1}{|\mathcal B|} \sum_{\mathbf x \in \mathcal B} p_i(\mathbf x).
\]
Grouping the terms in \eqref{eq:z_definition} by expert index:
\begin{equation}
Z = \sum_{i=1}^{N} f_i \cdot P_i.
\label{eq:z_fi_pi}
\end{equation}

Combining \eqref{eq:z_bound} and \eqref{eq:z_fi_pi}, we have:
\begin{equation}
\sum_{i=1}^{N} f_i \cdot P_i \ge \frac{1}{N}.
\label{eq:fi_pi_bound}
\end{equation}
Multiplying both sides by \(N\), we get the desired lower bound:
\begin{equation}
\mathcal L_{\text{load balancing}} = N \cdot \sum_{i=1}^{N} f_i \cdot P_i \ge 1.
\label{eq:final_lb}
\end{equation}

\textbf{Tightness.}
Equality holds when each sample has a uniform routing probability vector:
\[
p_1(\mathbf x) = \dots = p_N(\mathbf x) = \frac{1}{N}.
\]
In this case, \(f_i = P_i = \tfrac{1}{N}\) for all \(i\), and:
\[
\mathcal L_{\text{load balancing}} = N \cdot \sum_{i=1}^{N} \frac{1}{N} \cdot \frac{1}{N} = 1.
\]
\end{proof}

\clearpage
\newpage
\section{More Details about Datasets}
\label{app:dataset}
\begin{figure}[!h]
    \centering
    \includegraphics[width=1\linewidth]{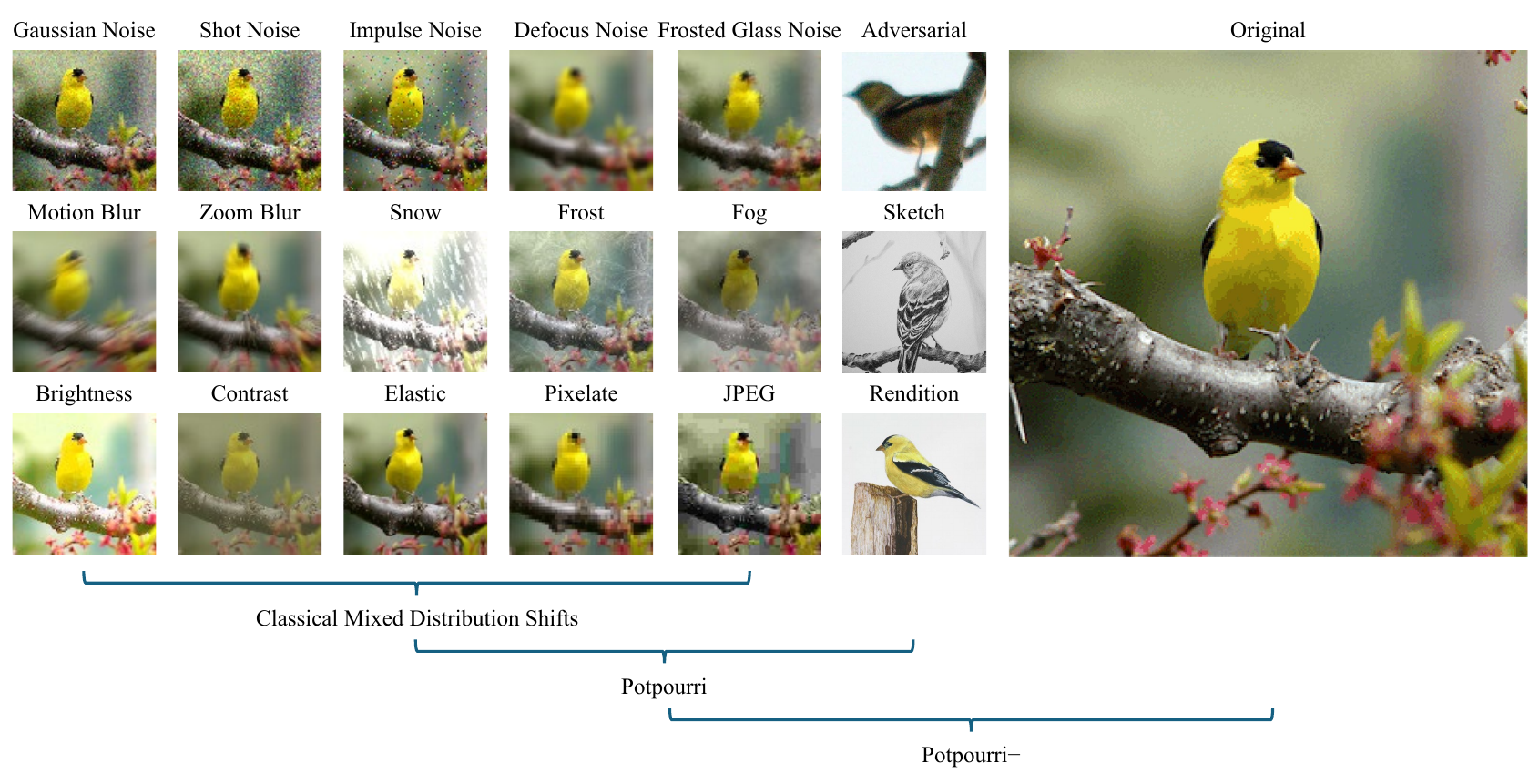}
    \caption{Overview of the corruption types and benchmark compositions used in our evaluation. The left block shows the 15 corruption types from ImageNet-C, grouped into four categories: noise, blur, weather, and digital. While classical mixed distribution shifts rely solely on these corruptions, our proposed Potpourri benchmark extends the evaluation to include samples from ImageNet-R (renditions), ImageNet-Sketch (sketches), and ImageNet-A (adversarial hard samples), thereby increasing semantic and stylistic diversity. Potpourri+ further includes clean validation images from the original ImageNet dataset to simulate real-world data streams that occasionally contain in-distribution (ID) samples.}
    \label{fig:benchmark}
\end{figure}
In this paper, we utilize the mixture of following datasets to validate the robustness of our methods against mixed distribution shifts setting: ImageNet-C~\citep{imagenet-c}, ImageNet-R~\citep{imagenet-r}, ImageNet-Sketch~\citep{imagenet-sketch}, ImageNet-A~\citep{imagenet-a} and the validation set of ImageNet~\citep{imagenet}.

\begin{table}[h]
\centering
    {\fontsize{9pt}{11pt}\selectfont
\begin{tabular}{lcc}
\toprule
\textbf{Dataset} & \textbf{\#Classes} & \textbf{\#Samples} \\
\midrule
ImageNet (validation set)       & 1,000 & 50,000 \\
ImageNet-C                    & 1,000 & 750,000 per level \\
ImageNet-R                    & 200   & 30,000 \\
ImageNet-Sketch          & 1,000 & 50,889 \\
ImageNet-A                    & 200   & 7,500 \\
\bottomrule
\end{tabular}
}
\caption{Summary of datasets used in the mixed distribution shifts evaluation.}
\label{tab:dataset_summary}
\end{table}

\paragraph{ImageNet~\citep{imagenet}} is a large-scale image classification dataset containing over 1.2 million training images and 50,000 validation images across 1,000 categories\footnote{The ImageNet dataset is distributed under a custom license, available at \texttt{https://www.image-net.org/download.php}.} In our evaluation, we use the validation set as a representative iID dataset. Its inclusion allows us to assess whether test-time adaptation methods, when applied to out-of-distribution (OOD) data, suffer from catastrophic forgetting—i.e., a significant drop in performance on previously well-learned ID samples.

\paragraph{ImageNet-C~\citep{imagenet-c}}consists of 15 diverse corruption types applied to the validation images of ImageNet\footnote{The ImageNet-C dataset is distributed under CC BY 4.0 license.}. These include Gaussian noise, shot noise, impulse noise, defocus blur, glass blur, motion blur, zoom blur, snow, frost, fog, brightness, contrast, elastic transform, pixelate, and JPEG compression, indexed as 0–14 in our paper. Each corruption type is provided at 5 severity levels, with 50,000 samples (covering 1,000 categories) per level. While the sample set remains the same across all levels, higher levels correspond to more severe distribution shifts.

\paragraph{ImageNet-R~\citep{imagenet-r}} contains 30,000 images from 200 ImageNet classes, rendered in a wide range of artistic styles such as paintings, cartoons, embroidery, and graphics\footnote{The ImageNet-R dataset is distributed under MIT license.}. It is designed to evaluate model robustness to style variation and domain shifts.

\paragraph{ImageNet-Sketch~\citep{imagenet-sketch}} consists of 50,889 black-and-white sketch images covering the same 1,000 classes as ImageNet\footnote{The ImageNet-Sketch dataset is distributed under MIT license.}. The sketches are collected from human drawings and represent strong shifts in visual modality compared to natural images.

\paragraph{ImageNet-A~\citep{imagenet-a}} includes 7,500 natural images across 200 classes that are challenging for standard models\footnote{The ImageNet-A dataset is distributed under MIT license.}. These images are selected to be adversarial in nature—i.e., they are misclassified by high-accuracy ImageNet models while being recognizable to humans.

\clearpage
\newpage
\section{More Implementation Details}
\label{app:implementaion_details}
All pre-trained models used for test-time adaptation in this paper are ViT-Base models obtained from the \texttt{timm} repository~\citep{timm}.
To ensure reproducibility, we have released the specific model checkpoints used in our experiments.\footnote{\texttt{https://drive.google.com/file/d/1RIrWyGnU1c12lI8BPByEmWDQI6Z3Nobk/view?usp=sharing}}
Due to the non-negligible discrepancy between the performance of the ViT-Base pre-trained weight we adopted and those of the previous studies used, as well as the sensitivity of test time adaptation methods to learning rate, we thus performed grid search on learning rate for compared methods. Detailed results of the grid search are provided in Appendix~\ref{app:grid_search}. For other hyper-parameters, we followed prior studies when values were provided; otherwise, we adopted fair configurations—for example, by aligning the number of activated parameters across methods.

In Tables~\ref{tab:mixed_shift_benchmark}, we report error bars to reflect the variability across multiple runs. Specifically, each result is averaged over three independent runs with fixed random seeds: 42, 4242, and 424242. The error bars represent the \textit{unbiased} standard deviation computed from these runs under identical settings with different random seeds.

\paragraph{Evaluation Metric.}
We use \textit{accuracy} as the primary evaluation metric, defined as the percentage of correctly classified samples over all test samples. Formally, for $N$ test samples, accuracy is calculated as:
\[
\mathrm{Accuracy} = \frac{1}{N} \sum_{i=1}^{N} \mathbb{I} \left[ \hat{y}_i = y_i \right],
\]
where $\hat{y}_i$ and $y_i$ denote the predicted and ground-truth labels for the $i$-th sample, respectively.

Accuracy is a widely used metric in image classification tasks due to its simplicity and interpretability. It provides a direct measure of a model's correctness, which aligns with our goal of evaluating test-time adaptation performance under mixed distribution shifts.

\paragraph{MoETTA(Ours).}
We use SGD as the update rule, with a momentum of 0.9. When batch size equals 64, we set learning rate to 1e-3.
Unless otherwise specified, all the LayerNorm except for the first one in ViT are replaced by MoE-LayerNorm.
For learnable parameters,inspired by Tent\citep{Tent}, we only update the parameters of router and experts in MoE-LayerNorm. For the classical mixed distribution shifts setting, we set \(\lambda\) in Eq.~\eqref{eq:lb_alpha} to 0.2 and use 9 experts. For the potpourri and potpourri+ settings, we set \(\lambda = 0.5\) and use 11 experts.

\paragraph{Tent\citep{Tent}.}
Besides learning rate, we follow all the hyper-parameters setting mentioned in Tent. 
Specifically, we use SGD as the update rule, and only set the affine parameters of normalization layers as trainable parameters.
According to the results of the grid search, when batch size equals 64, we set the learning rate to 5e-4.

\paragraph{EATA\citep{EATA}.}
Besides learning rate, we follow all the hyper-parameters setting mentioned in EATA. 
Specifically, the entropy constant $E_0$ for reliable sample identification is set to $0.4\times\ln 1000$. The $\epsilon$ for redundant sample identification is set to 0.05. The trade-off parameter $\beta$, regularization loss is set to 2000.
The number of pre-collected ID test samples for Fisher importance calculation is set to 2000. 
The update rule is SGD, with a momentum of 0.9. 
Only affine parameters of normalization layers are trainable.
According to the results of the grid search, when batch size equals 64, we set the learning rate to 6e-4.

\paragraph{CoTTA\citep{COTTA}.}
Besides learning rate, we generally follow all the hyper-parameters setting mentioned in CoTTA.
Specifically, we use SGD with a momentum of 0.9 as the optimizer, with an augmentation threshold $p_{th}$ of 0.1. If images are below this threshold, we use 32 augmentations. The restoration probability is set to 0.001, and the EMA factor $\alpha$ for teacher update is set to 0.999.
According to the results of the grid search, when batch size equals 64, we set the learning rate to 1e-3.

\paragraph{SAR\citep{SAR}.}
Besides learning rate, we follow all the hyper-parameters setting mentioned in SAR. 
Specifically, the entropy threshold,parameter $\rho$, moving average factor $e_m$ and reset threshold $e_0$ are set to $0.4\times\ln1000$, 0.05, 0.9, 0.2 respectively.
The update rule is SGD, with a momentum of 0.9. 
Only affine parameters of LayerNorm in blocks 0-8 are trainable.
According to the results of the grid search, when batch size equals 64, we set the learning rate to 5e-4.

\paragraph{DeYO\citep{DeYO}.}
Besides learning rate, we follow all the hyper-parameters setting mentioned in DeYO. 
Specifically, we set the parameter $Ent_0$ and $\tau_{Ent}$ to $0.4\times\ln1000$ and $0.5\times\ln1000$ respectively.
$\tau_{PLPD}$ is set to 0.2.
The update rule is SGD, with a momentum of 0.9. 
Only affine parameters of normalization layers are trainable.
According to the results of the grid search, when batch size equals 64, we set the learning rate to 5e-5.


\paragraph{BECoTTA\citep{BECoTTA}}
For a fair comparison, we use BECoTTA without SDA initialization. MoDE is injected into every block of the ViT backbone, with six experts per block. The expert rank is set to 1, as higher values were found to cause model collapse in mixed-domain scenarios. The number of domain routers is also set to 1. Based on the results of the grid search,  when the batch size equals 64, we set the learning rate to 1e-5.

\paragraph{MGTTA\citep{MGTTA}.}
Besides learning rate, we follow all the hyper-parameters setting mentioned in MGTTA. 
Specifically, we set the hyper-parameter $\lambda$ to 0.4,the GML hidden size to 8.
We adopted the pre-trained MGG released by the author.\footnote{\texttt{https://github.com/keikeiqi/MGTTA/blob/main/shared/mgg\_ckpt.pth}}
The update rule is SGD, with a momentum of 0.9. 
According to the results of the grid search, when batch size equals 64, we set the learning rate for $\phi^m$ to 3e-4.

\clearpage
\newpage
\section{Compute Resources}
\label{app:compute_resource}
All experiments were conducted on a Linux server (kernel version 5.4.0-176-generic) running Ubuntu. The machine is equipped with two Intel(R) Xeon(R) Gold 5318Y CPUs (96 cores), 251 GB of RAM, and two NVIDIA Tesla V100S GPUs with 32 GB of VRAM each. At the time of running the experiments, the NVIDIA driver version was 535.54.03, and the CUDA version was 12.2.
\clearpage
\newpage
\section{Visualization of the Cosine Similarity between Different Experts}
\label{app:cosine_similarity_visualization}
\begin{figure*}[!h]
    \centering
    \includegraphics{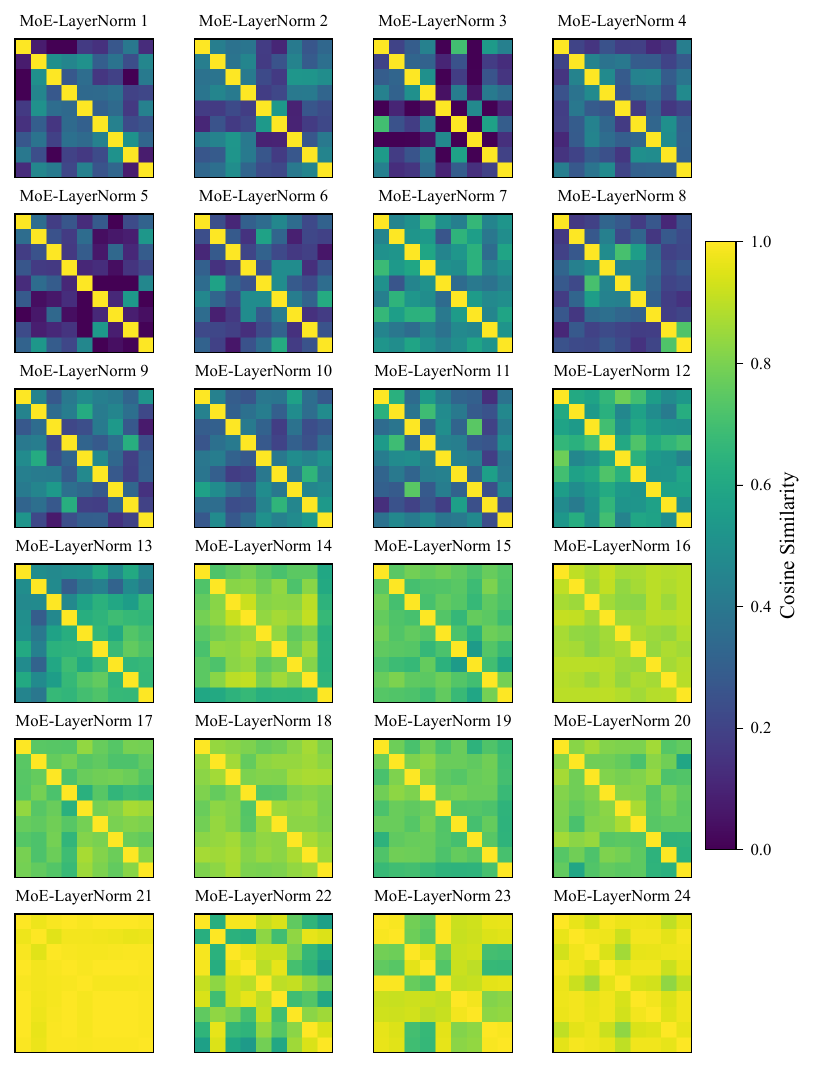}
\caption{Cosine similarity between the expert weights within each MoE-LayerNorm layer after adaptation under the potpourri+ setting with corruption level 5. Each heatmap corresponds to one MoE-LayerNorm layer.}
    \label{fig:weight_cosine_similarity}
\end{figure*}

\begin{figure*}[p]
    \centering
    \includegraphics{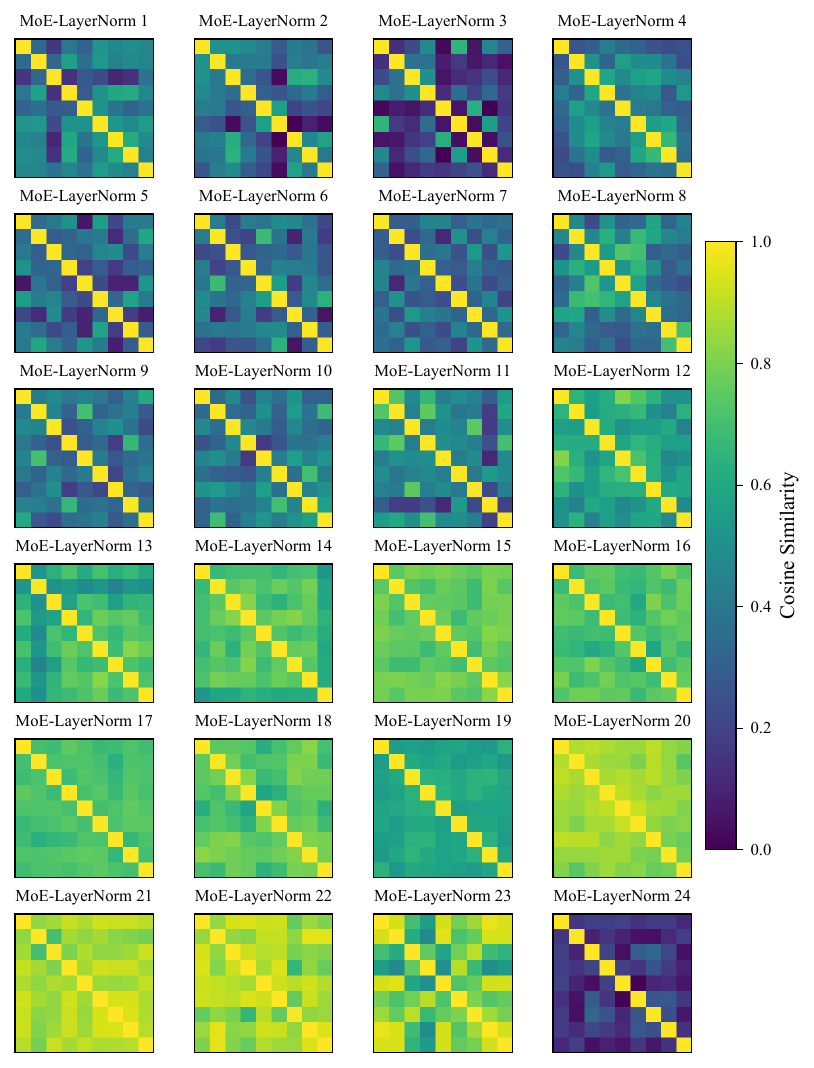}
\caption{Cosine similarity between the expert bias within each MoE-LayerNorm layer after adaptation under the potpourri+ setting with corruption level 5. Each heatmap corresponds to one MoE-LayerNorm layer.}
    \label{fig:bias_cosine_similarity}
\end{figure*}

We visualize the cosine similarity of experts in different MoE-LayerNorms after adaptation under the potpourri+ setting with corruption level 5. As shown in Fig.~\ref{fig:weight_cosine_similarity} and Fig.~\ref{fig:bias_cosine_similarity}, shallow MoE-LayerNorms exhibit greater variance among experts, which may be due to their role in handling texture features related to corruption.

\clearpage
\newpage
\section{Grid Search Results for Compared Methods}
\label{app:grid_search}

Since TTA methods are sensitive to the learning rate, for a fair comparison, we perform a grid search for all baselines using classical mixed distribution shifts at corruption level 5. The results are reported in Tab.~\ref{grid_search_Tent} to Tab.~\ref{grid_search_BECoTTA}.

\begin{table*}[ht]
  \centering
  \begin{tabular}{lccc}\\ 
    Learning rate     & 1e-4 & 5e-4 & 1e-3 \\ 
    \midrule
    Acc.(\%) & \underline{61.561}  & \textbf{63.139} & 53.619 \\ 
  \end{tabular}
    \caption{Grid search results on learning rate for Tent with classical mixed distribution shifts setting of ImageNet-C(level 5).}
  \label{grid_search_Tent}
\end{table*}

\begin{table*}[ht]
  \centering
  \begin{tabular}{lcccccccc}\\ 
    Learning rate     & 5e-4    & 6e-4           & 7e-4   & 8e-4   & 9e-4     & 1e-3   & 1.1e-3  & 1.2e-3 \\ 
    \midrule
    Acc.(\%)          & 62.213  & \textbf{64.178} & \underline{63.921} & 63.892 & 63.569   &  63.407  & 63.238  & 62.639  \\ 
  \end{tabular}
    \caption{Grid search results on learning rate for EATA with classical mixed distribution shifts setting of ImageNet-C(level 5).}
  \label{grid_search_EATA}
\end{table*}

\begin{table*}[ht]
  \centering
  \begin{tabular}{lccccc}\\ 
    Learning rate     & 5e-4 & 1e-3 & 5e-3 & 1e-2 & 5e-2   \\ 
    \midrule
    Acc.(\%) & \underline{58.689}  & \textbf{60.273} & 46.530 & 32.058 & 0.243 \\ 
  \end{tabular}
    \caption{Grid search results on learning rate for CoTTA with classical mixed distribution shifts setting of ImageNet-C(level 5).}
  \label{grid_search_CoTTA}
\end{table*}

\begin{table*}[!h]
  \centering
  \begin{tabular}{lccccccccccc}\\ 
    Learning rate  & 2e-4    & 3e-4    & 4e-4   & 5e-4   & 6e-4   & 7e-4    & 8e-4   &  9e-4  & 1e-3 & 1.1e-3 & 1.2e-3\\ 
    \midrule
    Acc.(\%)       & 60.396  & 60.599  & 60.671 & \textbf{60.760} & \underline{60.746} & 60.712   & 60.725 & 60.712 & 60.637  & 60.598  & 60.676 \\ 
  \end{tabular}
    \caption{Grid search results on learning rate for SAR with classical mixed distribution shifts setting of ImageNet-C(level 5).}
  \label{grid_search_SAR}
\end{table*}

\begin{table*}[!h]
  \centering
  \begin{tabular}{lcccc}\\ 
    Learning rate     & 1e-5 & 5e-5 & 1e-4 & 5e-4   \\ 
    \midrule
    Acc.(\%) & 62.451  & \textbf{63.931} & \underline{63.111} & 34.070 \\ 
  \end{tabular}
    \caption{Grid search results on learning rate for DeYO with classical mixed distribution shifts setting of ImageNet-C(level 5).}
  \label{grid_search_DeYO}
\end{table*}

\begin{table*}[!h]
  \centering
    {\fontsize{9pt}{11pt}\selectfont
  \begin{tabular}{lcccccccccccc}
    Learning rate     & 1e-4 & 2e-4 & 3e-4 & 4e-4 & 5e-4  & 6e-4 & 7e-4 & 8e-4 & 9e-4 & 1e-3 & 1.1e-3 & 1.2e-3 \\
    \midrule
    Acc.(\%) & 65.897  & \underline{66.195} & \textbf{66.210} & 66.048 & 65.882 & 65.643 & 65.395 & 65.162 & 64.689 & 63.929 & 63.185 & 61.159    \\
  \end{tabular}
  }
    \caption{Grid search results on learning rate for MGTTA with classical mixed distribution shifts setting of ImageNet-C (level 5).}
  \label{grid_search_MGTTA}
\end{table*}

\begin{table*}[!h]
  \centering
    {\fontsize{9pt}{11pt}\selectfont
  \begin{tabular}{lcccccccccccc}
    Learning rate     & 1e-5 & 2e-5 & 3e-5 & 4e-5 & 5e-5  & 1e-4 & 5e-4 \\
    \midrule
    Acc.(\%) & \textbf{61.563}  & \underline{46.187} & 34.417 & 22.744 & 19.466 & 11.168 & 1.960 \\
  \end{tabular}
  }
    \caption{Grid search results on learning rate for BECoTTA with classical mixed distribution shifts setting of ImageNet-C (level 5).}
  \label{grid_search_BECoTTA}
\end{table*}
\clearpage
\newpage
\section{Scalability}
\label{app:scalability}
\begin{table}[h]
  \centering
  {\fontsize{9pt}{11pt}\selectfont
  \begin{tabular}{>{\centering\arraybackslash}c c cccccccc c}
    \toprule
    Model & Setting & Noadapt & Tent & EATA & CoTTA & SAR & DeYO & MGTTA & BECoTTA & Ours \\
    \midrule
    \multirow{3}{*}{\makecell{ViT \\ -L/16}} 
        & Classical & 61.0 & 64.7 & \underline{66.6} & - & 65.5 & 65.6 & 61.7 & 66.2 & \textbf{68.8} \\
        & Pot. & 59.8 & 62.4 & \textbf{64.6} & - & 62.0 & 63.6 & 60.6 & 1.5 & \underline{64.4} \\
        & Pot.+ &61.2 & 65.4 & 1.4 & - & 60.5 & 65.3 & 61.6 & \underline{66.0} & \textbf{68.0} \\
    \bottomrule
  \end{tabular}
    }
    \caption{
    Accuracy comparison (\%, $\uparrow$) under different mixed distribution settings using ViT-L/16. Results for CoTTA are not reported due to out of memory error. The best performance is highlighted in \textbf{bold} and the second best is indicated by \underline{underlining}. This convention is followed in all subsequent tables.
  }
\end{table}
\clearpage
\newpage
\section{Broader Impacts}
\label{app:broader_impacts}
Our work focuses on improving test-time adaptation under mixed distribution shifts, a scenario commonly encountered in real-world deployments. By enabling models to adapt dynamically to heterogeneous test samples without requiring labeled data, our method has the potential to improve the robustness and reliability of AI systems in high-stakes applications such as autonomous driving, medical diagnosis, and environmental monitoring. For example, our approach can help maintain consistent model performance across weather conditions or medical imaging devices from different institutions. These capabilities contribute toward safer and more generalizable AI deployment in practice.

We are not aware of any direct negative societal impacts associated with this work.

\end{document}